\documentclass{article}

\usepackage{microtype}
\usepackage{graphicx}
\usepackage{xurl}
\usepackage{hyperref}

\usepackage{subfigure}
\usepackage{booktabs} %
\usepackage{multirow}
\usepackage{hyperref}

\usepackage{amsmath}
\usepackage{amssymb}
\usepackage{amsfonts}
\usepackage{bm}
\usepackage{tcolorbox}
\tcbuselibrary{skins}

\usepackage[table]{xcolor}

\definecolor{myRed}{HTML}{ED3F4B}
\definecolor{myOrange}{HTML}{F9973D}
\definecolor{myGreen}{HTML}{00BA00}

\tcbset{
  aibox/.style={
    top=10pt,
    left=5pt,
    colback=white,
    colframe=black,
    colbacktitle=black,
    enhanced,
    center,
    attach boxed title to top left={yshift=-0.1in,xshift=0.15in},
    boxed title style={boxrule=0pt,colframe=white,},
  }
}
\newtcolorbox{AIbox}[3][]{aibox,title=#2,#1,width=#3}

\usepackage[accepted]{icml2026}

\usepackage{amsmath}
\usepackage{amssymb}
\usepackage{mathtools}
\usepackage{amsthm}
\usepackage{xspace}
\usepackage[capitalize,noabbrev]{cleveref}
\definecolor{darkgreen}{RGB}{0, 100, 0} %
\theoremstyle{plain}

\theoremstyle{definition}

\theoremstyle{remark}

\newcommand{\method}{Distribution Aligned Imitation Learning\xspace}

\usepackage[textsize=tiny]{todonotes}

\icmltitlerunning{Making Expert Reasoning Learnable with Self-Distillation}

\begin{document}

\twocolumn[
\icmltitle{Making Expert Reasoning Learnable with Self-Distillation}

\icmlsetsymbol{equal}{*}

\begin{icmlauthorlist}
\icmlauthor{Ethan Mendes}{yyy}
\icmlauthor{Jungsoo Park}{yyy}
\icmlauthor{Alan Ritter}{yyy}
\end{icmlauthorlist}

\icmlaffiliation{yyy}{Georgia Institute of Technology, Atlanta, Georgia}

\icmlcorrespondingauthor{Ethan Mendes}{emendes3@gatech.edu}

\icmlkeywords{Machine Learning, ICML}

\vskip 0.3in
]

\printAffiliationsAndNotice{} %

\begin{abstract}

Improving the reasoning capabilities of large language models (LLMs) typically relies either on the model's ability to sample a correct solution to be reinforced or the existence of a stronger model able to solve the problem. However, many difficult problems remain intractable for even current frontier models, preventing the extraction of valid training signals. A promising alternative is to leverage high-quality expert human solutions, yet naive imitation of this data fails because it is fundamentally out-of-distribution: expert solutions are typically \textit{didactic}, containing implicit reasoning gaps intended for human readers rather than computational models. Furthermore, high-quality expert solutions are expensive, necessitating generalizable, sample-efficient training methods. We propose Distribution Aligned Imitation Learning (DAIL), a two-step self-distillation method that bridges the distributional gap by first transforming expert solutions into detailed, in-distribution reasoning traces and then applying a contrastive objective to focus learning on expert insights and methodologies. We find that DAIL can leverage fewer than 1000 high-quality expert solutions to achieve up to 31\% pass@$128$ gains on Qwen2.5-Instruct and Qwen3, double reasoning efficiency, and enable out-of-domain generalization. 
\end{abstract}

\section{Introduction}
\label{sec:introduction}

Large reasoning models (LRMs)~\cite{yang2025qwen3, guo2025deepseek, openai_o3_o4mini_2025, google2025gemini3}
have recently demonstrated an impressive ability to solve extremely challenging tasks, from competition mathematics problems to graduate student exam questions~\cite{rein2024gpqa}.
These models acquire this reasoning ability primarily through a stage of reinforcement learning with verifiable rewards (RLVR)~\cite{shao2024deepseekmath}, an online RL procedure where models are rewarded based on the correctness of their final answer ($\hat{y} =y$) on large datasets of reasoning questions. 

However, during RLVR, a problem contributes to learning only if the model can sample a rollout that produces the correct answer.
Paradoxically, on the most difficult problems, from which we would like them to learn the most, models often fail to extract any training signal as the rewards, advantages, and gradients are 0~\cite{nan2025ngrpo, chen2025nudging, chen2025spectral}. To bypass this exploration bottleneck, models could instead learn from high-quality
human expert reasoning traces, such as those seen in recent evaluation benchmarks~\cite{glazer2024frontiermath, phan2025humanity}.
However, directly fine-tuning a post-trained model on input-output pairs $\mathcal{D} = \{(x_i,s_i)\}_{i = 1}^n$ to imitate expert solutions $s$ leads to a severe reasoning performance deterioration~\cite{yang2026int}. This degradation appears to arise because expert human targets are fundamentally out-of-distribution (OOD) from the model's own reasoning process learned during post-training. Specifically, expert solutions are typically \textit{didactic}; they prioritize human readability by omitting granular intermediate steps that, while implied for a human reader, are essential for the model to reason successfully. Additionally, human solutions lack explicit search dynamics, such as backtracking and self-correction, that characterize some long chain-of-thought (CoT) LRM generations~\cite{marjanovic2025deepseek} after training with RLVR. Consequently, standard behavioral cloning forces the model to \textit{shortcut} its internal reasoning process acquired during post-training, collapsing performance.

\begin{figure*}[t]
    \centering
    \includegraphics[width=0.96\linewidth]{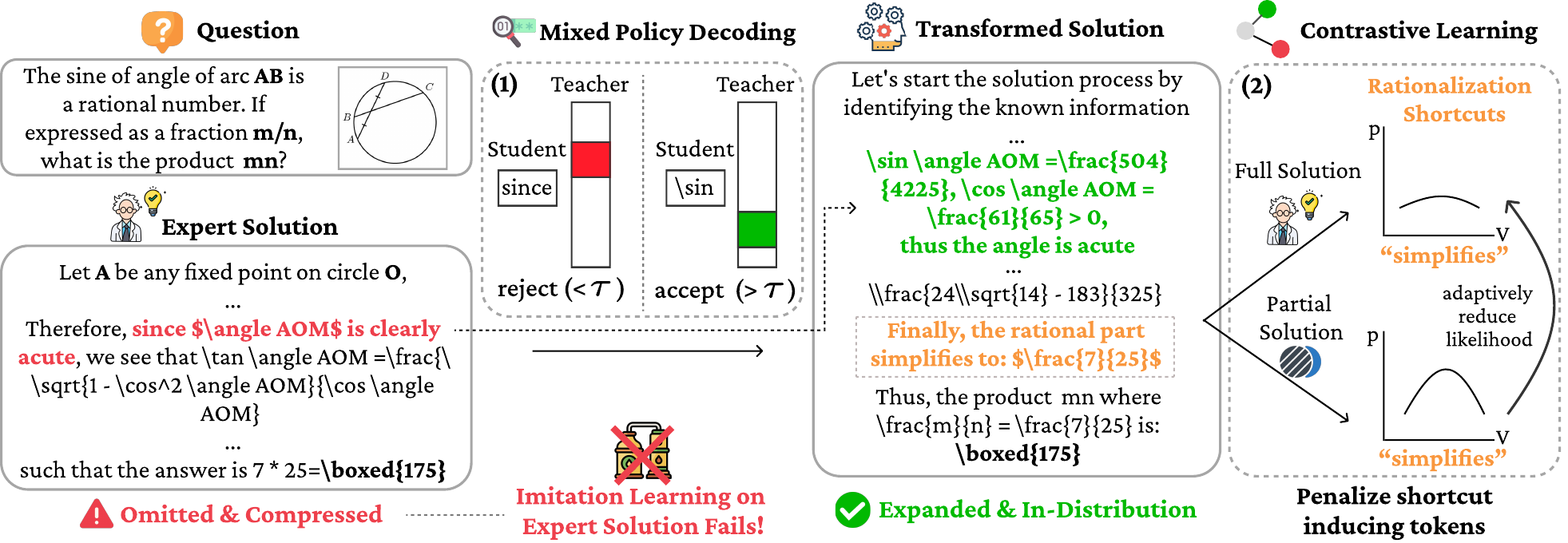}
    \caption{\textbf{Overview of DAIL}. Starting from a small set of expert solutions, we generate in-distribution transformed solutions for training via \emph{mixed policy decoding}: the student model uses the expert solution as a reference to produce a detailed reasoning trace. This process mitigates \textit{didactic shortcuts}, e.g., the expert solution skips the proof of {why $\angle AOM$ is acute} (in \textcolor{myRed}{\textbf{red}}), while the transformed solution {explicitly details this reasoning} (in \textcolor{myGreen}{\textbf{green}}). Using this dataset of transformed solutions, DAIL applies contrastive learning on paired {full} vs. {partial} solutions to discourage imitating \textit{rationalization shortcuts} in transformed solutions, e.g., the model ignoring the irrational part of a derived result to {force the generation of the correct answer} (in \textcolor{myOrange}{\textbf{orange}}).} 
    \label{fig:concept_figure}
\end{figure*}

In this paper, we propose \textbf{\method(DAIL)}, a novel post-training method that enables a student model to learn directly from high-quality expert solutions to complex problems (see Figure~\ref{fig:concept_figure}). 
Because collecting this data requires domain experts and could cost upwards of \textit{\$1,000 per sample}~\cite{chiou2025inside}, $\mathcal{D}$ is practically small. In this regime, we aim to maximally leverage each high-cost instance for generalizable reasoning improvement.
DAIL bridges the mentioned distribution gap between expert solutions and a student's own reasoning process by transforming highly OOD expert solutions into learnable training examples. Specifically, we propose a mixed policy rollout process where the student generates reasoning traces in tandem with a ``teacher'', a frozen instance of the model conditioned with the ground-truth solution.
This process expands the often non-comprehensive expert solution into a detailed, in-distribution reasoning chain and fills in \textit{didactic shortcuts} in the human-written solution, such as skipped steps and implied calculations. However, the cost of this distributional alignment is that generated traces may contain \textit{rationalization shortcuts} or missing or insufficient justifications that lead to an intermediate result in the expert's solution.
As standard negative log-likelihood (NLL) loss enforces indiscriminate token-level imitation, it compels the student to internalize these harmful shortcuts. We find this naive mimicry ultimately degrades the student's generalizable reasoning ability.
Instead, we propose a contrastive objective designed to prevent the student model from learning superficial reasoning. This objective penalizes mimicking a ``negative reference'' conditioned only on isolated intermediate results, enabling robust reasoning improvements.

Building on findings~\cite{zhou2023lima, muennighoff2025s1} that fine-tuning on even a limited number of high-quality examples can yield substantial performance gains, we demonstrate that applying DAIL on a dataset of fewer than 1000 expert human solutions leads to significant improvements across two settings. For Qwen2.5-Instruct, a non-reasoning, post-trained model, we utilize a dataset of competition mathematics problems that the model itself failed to solve, even with repeated sampling (i.e., for high $k$, pass@$k$ = 0). To train Qwen3 (think), an LRM, we collect a novel dataset of Olympiad proofs authored by a current International Math Olympiad coach, which we publicly release as a new resource.\footnote{We were given permission to use and release this dataset.} This latter setting highlights that DAIL enables learning on non-verifiable proof problems, which is not possible with standard RLVR without a generative reward model. Across both settings, we find substantial improvements in pass@$k$ on challenging mathematics reasoning benchmarks such as AIME 2024/2025, BeyondAIME~\cite{seed2025seed1}, and IMO-AnswerBench~\cite{luong2025towards}. Furthermore, for LRMs, we find that performance improvements due to DAIL scale well as model size and test-time compute are increased. To our knowledge, this is the first demonstration of improving the reasoning ability of long-CoT reasoning while learning directly from expert solutions. 
We also find that DAIL generalizes well to the out-of-domain GPQA-diamond benchmark~\cite{rein2024gpqa}, while also promoting \textit{efficient reasoning}, as it matches or exceeds the performance of untrained models with $2\times$ fewer tokens.

\section{\method}

In this section, we outline DAIL, an offline learning method that improves a student model $M_\theta$ on a dataset $\mathcal{D} = \{(x_i,s_i)\}^n_{i=1}$ consisting of high-quality expert solutions $s$ to complex, potentially non-verifiable, problems $x$.

Directly training on such data via standard behavior cloning (BC) fails due to a fundamental distribution mismatch between expert human reasoning and the student's reasoning. DAIL addresses this via a two-stage process. First, we bridge this distribution gap by synthesizing expanded reasoning traces that are in-distribution for $M_\theta$ yet grounded in the expert solutions (\S\ref{subsec:generating_learnable_reasoning}). This approach resolves \textbf{didactic shortcuts} (see the left side of Figure~\ref{fig:concept_figure}) in expert traces (e.g., ``solving with the quadratic formula, we obtain...'') that omit the granular reasoning steps required by the student to reason.
Second, we optimize a contrastive objective (\S\ref{subsec:learning}) to mitigate the side-effects of this in-distribution expansion.

\subsection{Generating In-Distribution Learnable Reasoning}
\label{subsec:generating_learnable_reasoning}

For any pair $(x, s)$, we define the \textit{teacher} $M_\text{T}(\cdot) = M_{\theta_{\text{ref}}}(\cdot|x, s)$. Here, $\theta_{\text{ref}}$ denotes the frozen initial weights of the student model prior to optimization. The goal of this generation step is to use the teacher to generate a version $r$ of each expert solution $s$ that is in-distribution for the student $M_\theta(\cdot|x)$, resulting in a new synthetic dataset $\mathcal{D}_{\text{syn}} = \{(x_i, r_i)\}_{i=1}^n$.

\paragraph{Direct Sampling.}
A simple way to generate $\mathcal{D}_{\text{syn}}$ is to sample solution trajectories $r$ directly from the teacher, i.e., $r_i \sim M_{\text{T}}(\cdot| r_{<i}) = M_{\theta_{{\text{ref}}}}(\cdot|x, s, r_{<i})$. We find that this approach is surprisingly effective for non-reasoning instruction models like Qwen2.5-Instruct, if we carefully prompt the teacher during this generation process to fill in reasoning gaps in $s$, and solve the problem as if it were solving it without guidance (see the full prompt in Table~\ref{fig:teacher_prompt}).

\paragraph{Mixed Policy Rollouts.}
However, generating high-quality reasoning traces via direct sampling on LRMs like Qwen3 (think) is challenging due to the reflective nature of these models. Specifically, traces generated with access to the expert solution via direct sampling often contain references to this solution despite explicit prompting. Additionally, these traces exhibit fewer self-correction steps because the strong guidance of the expert solution allows the generation process to bypass the model's natural verification mechanisms, resulting in less authentic reasoning. 

To mitigate these effects, we propose a mixed policy rollout approach. Inspired by speculative decoding~\cite{leviathan2023fast} and interactive imitation~\cite{ross2011reduction}, our approach has the student and the teacher work in tandem to generate reasoning traces, with the student deferring to the teacher only when necessary. Specifically, the $i$th token of a generation $r$ is formed by first sampling from the student, i.e., $t \sim M_\theta(\cdot|x, r_{<i})$, and verifying it with the teacher, i.e., confirming $M_\text{T}(t|r_{<i}) \ge \tau$, for a fixed hyperparameter $0 \le  \tau \le 1$. If this verification is successful, $r_i := t$, and otherwise, we sample from the teacher: $r_i \sim M_\text{T}(\cdot|r_{<i})$. 

Unlike work on improving traditional distillation~\cite{xu2024speculative}, which jointly generates with the student and a larger teacher model to ensure the teacher can provide valuable guidance to the student, mixed policy rollouts aim to ensure the resulting traces preserve the student's natural reasoning flow and authentic artifacts, while utilizing the teacher mainly to anchor generation to the expert solution.

\subsection{Learning from Expert-Grounded Traces}
\label{subsec:learning}
After generating the synthetic dataset $\mathcal{D}_{\text{syn}}$, the natural next step is to optimize the student model $M_\theta$ using these expanded reasoning traces. A naive approach is to use BC via NLL loss, i.e., minimize $\mathcal{L}_{\text{NLL}} (\theta) = - \mathbb{E}_{(x,r) \sim \mathcal{D}_{\text{syn}}} [\log M_\theta(r|x)]$. 

However, NLL is ill-suited for training on $\mathcal{D}_{\text{syn}}$ due to the presence of \textbf{rationalization shortcuts} in the expanded reasoning traces.  Unlike didactic shortcuts, which are logical omissions made by human experts for brevity, rationalization shortcuts are deficient logical bridges to an intermediate result contained in the expert solution. Specifically, because the model generating the trace has access to the expert solution $s$ in its context, it often forces a mathematical derivation to reach the known result rather than strictly deriving it (see the right side of Figure~\ref{fig:concept_figure}). %
Standard NLL indiscriminately forces the student to imitate every token in the trace with equal weight. This applies regardless of whether a token represents a valid logical progression or a spurious shortcut. Consequently, models trained via NLL tend to memorize these flawed heuristics, leading to poor generalization when the expert solution is unavailable at inference time.

\paragraph{Mitigating Learning Rationalization Shortcuts.}

To mitigate this behavior, we introduce a contrastive objective designed to penalize the imitation of shortcuts, thereby improving the model's generalization to more complex, out-of-distribution tasks. 
This objective explicitly addresses the known limitations of BC on suboptimal datasets, where models tend to indiscriminately imitate both effective reasoning and spurious shortcuts~\cite{kumar2022should}.

Concretely, we construct a \textit{negative reference} model that preferentially produces rationalized shortcut-laden reasoning. Like the teacher, the negative reference shares the student’s parameters but is conditioned differently. Specifically, this model is defined as $M_\text{NR}(\cdot) = M_{\theta_{\text{ref}}}(\cdot|x, \Tilde{s})$, where $\Tilde{s}$ is a partial solution to $x$ obtained by deleting intermediate reasoning components of $s$ and retaining only coarse-grained solution waypoints. We construct $\tilde{s}$  automatically using a regular expression (see Appendix~\ref{appendix:dataset_gen}). For the math domain, $\tilde{s}$ consists of a list of key numerical and symbolic results from the expert solution. Conditioning on these waypoints shifts the model's probability distribution to favor bypassing step-by-step logical progression and jumping directly between intermediate results, effectively characterizing the shortcut behavior we aim to penalize.

Using this negative reference model, we can optimize the following contrastive loss function:

\begin{equation*} \label{eq:loss}
\resizebox{\hsize}{!}{$
\begin{aligned}
\mathcal{L}(\theta) = \mathbb{E}_{(x,r) \sim \mathcal{D_\text{syn}}} \Bigg[ \sum_{t=1}^{|r|} \Big( 
&D_{\text{KL}}\big(M_\theta(\cdot|x, r_{<t}) \parallel M_\text{T}(\cdot|r_{<t})\big) \\
- \gamma &D_{\text{KL}}\big(M_\theta(\cdot|x, r_{<t}) \parallel M_\text{NR}(\cdot|r_{<t})\big) \Big) \Bigg]
\end{aligned}
$}
\end{equation*}

where $D_{\text{KL}}$ is the token-level KL divergence and $0 \le \gamma \le 1$ is a hyperparameter. 
The objective effectively reduces the likelihood of tokens that have a high probability under the ``negative reference'' relative to the full-information teacher. 
While maximizing divergence via the negative term is theoretically unbounded, we find that because the student is initialized with the same weights as the teacher and negative reference and strictly anchored by the positive distillation term, training remains stable (see Appendix~\ref{appendix: compute}).

\paragraph{Training Efficiency.}
The above objective provides an off-policy, offline training framework that offers two primary efficiency advantages: 

\textbf{(1) Asynchronous Data Generation:} Unlike online RLVR approaches~\cite{shao2024deepseekmath,guo2025deepseek} that require interleaved generation and optimization, the in-distribution generation process is entirely decoupled from the training loop. This allows for massive parallelization of the dataset construction across distributed clusters before optimization.

\textbf{(2) Efficient Memory Management:} Since $M_{\theta_{\text{ref}}}$ and $M_{\theta}$ share identical base parameters at the start of training, we can represent the student updates using a LoRA adapter~\cite{hu2022lora}. By selectively enabling the adapter, we can compute the forward passes for the teacher and negative reference using the same frozen model $\theta_{\text{ref}}$. Thus, we only need to store a single copy of the model weights in memory, significantly reducing hardware requirements compared to traditional distillation approaches.

\section{Results}

\begin{figure*}[t]
\centering

\includegraphics[width=\linewidth]{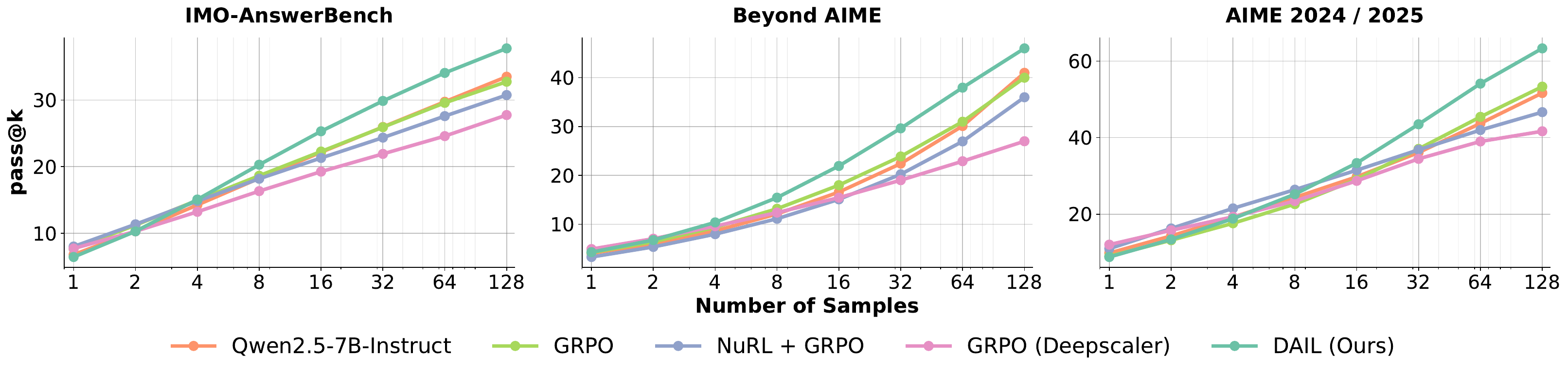}
\caption{Pass@$k$ performance comparison of DAIL on Qwen2.5-7B-Instruct with \texttt{e1-verifiable} compared to RLVR methods on IMO-Answer, BeyondAIME, and AIME 2024 / 2025 benchmarks. DAIL exhibits consistent performance improvements over the base instruction model, while applying RLVR methods results in pass@k reductions due to the difficulty of training dataset problems.}
\label{fig:qwen_2.5_results}

\end{figure*}

\begin{figure}[t]
    \centering
    \includegraphics[width=\linewidth]{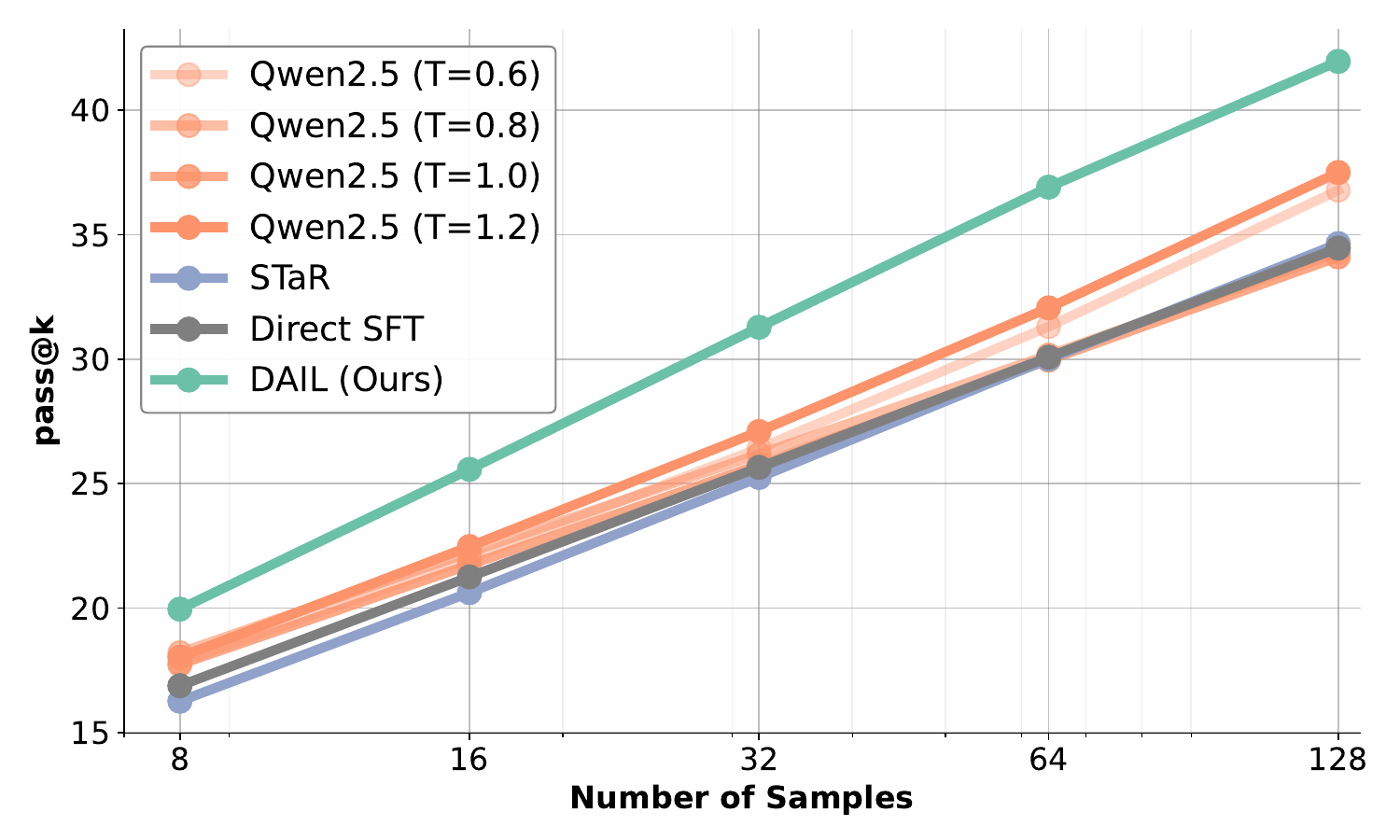}
    \caption{\textbf{Baselines.} Comparing pass@k performance of DAIL to temperature, STaR rationalization, and direct SFT on expert solutions baselines. To better visualize the relative performance between baselines, the results are plotted for $k \ge 8$. See Figure~\ref{fig:all_naive_baselines} for results at lower $k$ values.}
    \label{fig:naive_baselines}
\end{figure}

\subsection{Experimental Setup}
\label{subsec:setup}
\paragraph{Curated Training Datasets.}
As mentioned briefly in \S\ref{sec:introduction}, we curate two datasets of reasoning problems for training. 

The first dataset \textbf{\texttt{e1-verifiable}} consists of $417$ problems and solutions from historical American Invitational Mathematics Examinations (AIME) from the years 1985 to 2023. To measure how well DAIL can enable learning from difficult problems, we only retain problems that were not solved by the target Qwen2.5-7B-Instruct model in 32 attempts. Since official solutions to past exams are difficult to find online, we utilize community-generated solutions collected from the AoPS Wiki~\footnote{\url{https://artofproblemsolving.com/wiki/index.php?title=AIME_Problems_and_Solutions}} by~\citet{yue2024harp}. This process yields a final set of 417 problems that constitute \texttt{e1-verifiable}. Training on this dataset allows for a direct comparison between DAIL and RLVR methods.

We also curate \textbf{\texttt{e1-proof}}, which consists of $669$ non-verifiable Olympiad proof problems from the International Mathematics Olympiad and other regional contests. Since these are open-ended proof problems, it is not possible to train on them using conventional correctness-based RLVR approaches. We sourced expert solutions to these problems from USA IMO Coach Evan Chen's website\footnote{\url{https://web.evanchen.cc/problems.html}} with permission.\footnote{Our extracted training dataset is available at \url{https://huggingface.co/datasets/emendes3/e1-proof}.}

Finally, we also demonstrate the efficacy of DAIL on non-math domains in Appendix~\ref{appendix:non_math}.

\paragraph{Training Configuration.}
We train Qwen2.5-7B-Instruct on \texttt{e1-verifiable} and Qwen3-8B (think) and Qwen3-14B (think) on \texttt{e1-proof}. As mentioned in \S\ref{subsec:generating_learnable_reasoning}, for Qwen3 models, we use mixed policy rollouts. We investigate the effect of mixed policy rollouts and other factors during generation and learning in \S\ref{subsec:ablations}. For more details about training, please see Appendix~\ref{appendix:train-gen-eval}.\footnote{Our code is available at \url{https://github.com/ethanm88/DAIL}}

\paragraph{Evaluation Datasets.}

To quantify how DAIL scales to problems more challenging than the training distribution, we evaluate performance on three mathematical reasoning benchmarks as well as an out-of-domain science reasoning benchmark to measure generalization:

\textit{AIME 2024 / 2025}: A collection of 60 problems from the two most recent editions of the American Invitational Mathematics Examination (AIME). All answers are integers within the range $[0, 999]$.

\textit{BeyondAIME}~\cite{seed2025seed1}: A verifiable benchmark of 100 problems designed to mirror the difficulty of the final five (hardest) problems of standard AIME exams. Unlike standard AIME, the integer answers are unbounded.

\textit{IMO-AnswerBench}~\cite{luong2025towards}: A curated dataset of  400 Olympiad-level problems across algebra, combinatorics, geometry, and number theory. These problems are manually rewritten by experts to be verifiable and resistant to memorization. 

\textit{GPQA-Diamond}~\cite{rein2024gpqa}: A collection of 198 graduate-level biology, physics, and chemistry multiple choice questions written by domain experts.

\paragraph{Evaluation Metrics.}
We use the pass@$k$ metric, which quantifies the probability of obtaining a correct sample when sampling $k$ times. This metric can be computed via the following unbiased estimator~\cite{chen2021evaluating}:
\begin{equation*}
    \text{pass}@k = \mathbb{E}_{x \sim\mathcal{D}_{\text{eval}}}\left[1 - \frac{{n - c \choose k}}{{n \choose k}}\right]
\end{equation*}
where $n \ge k$ is the number of samples generated by the model for each problem and $c \le n$ is the number of correct answers generated by the model for a specific problem. We set $n = 128$ and report results for $k \in \{1, 2, 4, \dots, 128\}$.

For difficult mathematics questions like those in BeyondAIME and IMO-AnswerBench, greedy accuracy (pass@$1$) is inherently low ($\approx 5\%$). Thus, the practical utility of these models depends on their ability to uncover correct solutions via repeated sampling. Given the demonstrated tradeoff between these metrics~\cite{yue2025does,tang2025optimizing}, we seek significant gains in search potential (pass@$k$ for $k \gg 1$) that do not come at a prohibitive cost to greedy stability (pass@$1$).

\begin{figure*}[t]
    \centering
    \includegraphics[width=\linewidth]{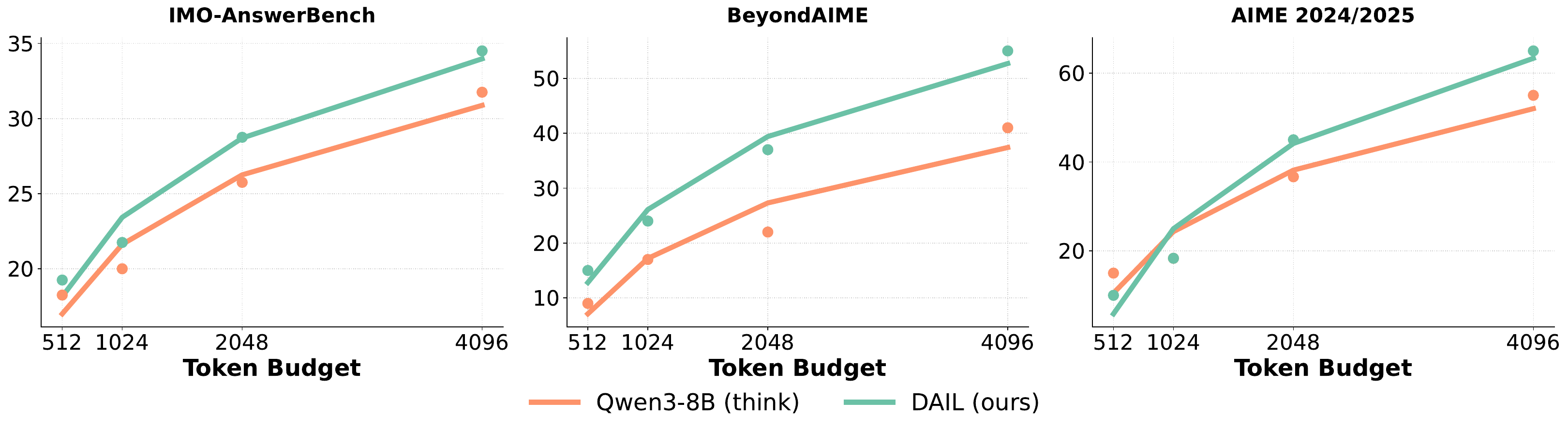}
    \includegraphics[width=\linewidth]{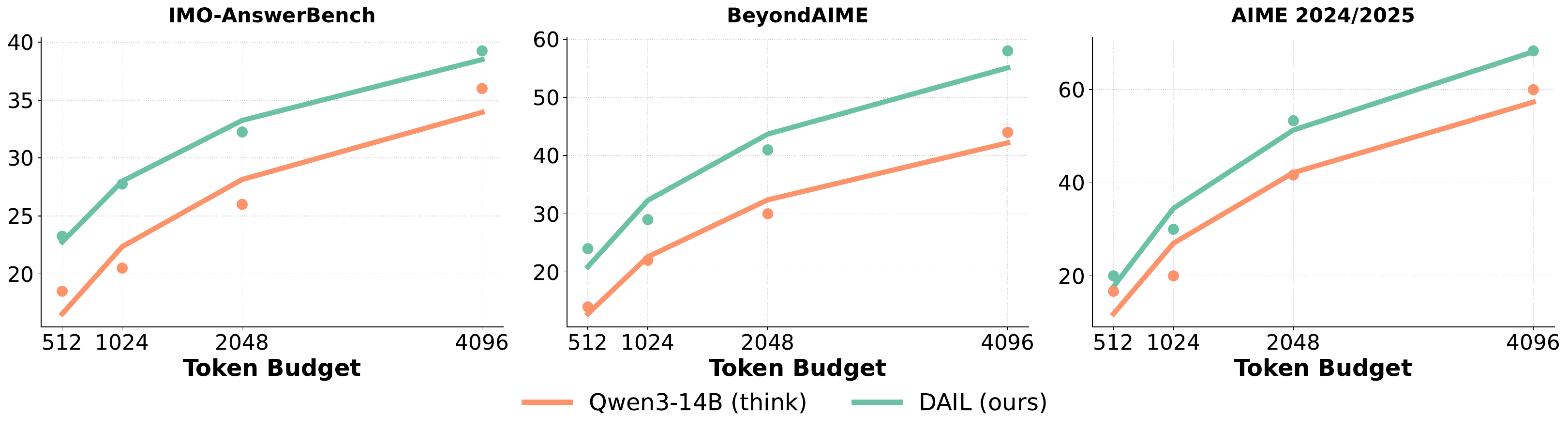}
    \vspace{-0.7cm}
    \caption{\textbf{Test-Time Efficiency.} Coverage (pass@128) on mathematics benchmarks under various token reasoning limits. To ensure gains are not due to non-response, models are prompted to provide a final answer after these token limits. Compared to Qwen3-8B (think) and Qwen3-14B (think), the models trained with DAIL on \texttt{e1-proof} yield improved performance across token budgets and benchmarks.}
    \label{fig:reasoning_model_efficiency}
\end{figure*}

\paragraph{Baselines.}
\label{subsec:baselines}
We evaluate the performance of DAIL against a comprehensive set of baseline methods:

\textit{Naive Baselines:}
We first compare DAIL to standard imitation learning via Supervised Fine-Tuning (SFT) on expert human solutions. We also compare to STaR Rationalization~\cite{zelikman2022star}, which induces the model to generate plausible solutions using the numerical answer as a hint; these rationalizations are subsequently used for SFT. Finally, to ensure improvements are not solely due to inference parameters, we evaluate performance across a temperature range of $\tau \in \{0.6, 0.8, 1.0, 1.2\}$, as variations in temperature are known to influence pass@$k$ performance~\cite{du2025optimizing, wu2025role}. Unless specified, we otherwise default to a temperature of 0.6.

\textit{Correctness-based RLVR:} We further benchmark against correctness-based RLVR methods trained on answer-verifiable tasks. Specifically, we compare against Group Relative Policy Optimization (GRPO)~\cite{shao2024deepseekmath} and NuRL~\cite{chen2025nudging}. The latter addresses the challenge of reward sparsity on hard problems by providing high-level hints derived from the ground truth when standard rollouts fail to yield a correct answer. Finally, we also compare to a realistic RLVR setup by training on the $>40$K example DeepScaleR dataset~\cite{deepscaler2025} with GRPO.

\subsection{DAIL Expands Math Reasoning Ability}
\label{subsec:improved_reasoning_ability}
In this section, we focus on performance improvements of DAIL when applied to Qwen-2.5-7B-Instruct by training on \texttt{e1-verifiable}. This verifiable problem setting allows us to rigorously evaluate and contextualize DAIL's improvements compared to baseline methods that may require reward.

\paragraph{DAIL enables robust generalization where RLVR fails.}
In Figure~\ref{fig:qwen_2.5_results}, we compare DAIL against the Qwen2.5-7B-Instruct baseline and standard RLVR approaches. DAIL consistently outperforms the base instruct model across all benchmarks, with the performance gap widening significantly at larger $k$ values. Moreover, DAIL enables consistent improvements on BeyondAIME and IMO-AnswerBench, which contain more difficult problems than \texttt{e1-verifiable}.%

Since \texttt{e1-verifiable} problems were chosen such that they were not solved by the base instruct model in 32 attempts, if these problems were truly unsolvable, GRPO should not update model parameters as rewards, advantages, and policy-gradient loss would be zero across training problems. However, practically, the model may sample a few correct rollouts on which it can take gradient steps. But, instead of generalizing, the model overfits to these rare, stochastic successes, leading to the observed degradation in general reasoning ability as evidenced by the downward shift in the pass@$k$ curve. We confirm this behavior in our analysis in \S\ref{subsec:ablations}. 

Most notably, GRPO on the $>40$K sample DeepScaleR dataset shows only marginal gains in pass@$1$ over the base instruct model on only a subset of the evaluated benchmarks. Moreover, this model shows a degradation in pass@$k$ at larger $k$ values, which is consistent with findings in prior work~\cite{yue2025does}. Therefore, simply scaling verifiable math data fails to enable Olympiad-level reasoning.

While the hint-based NuRL method has shown promise in a large-scale training regime, requiring $\approx$ 1k GPU hours for convergence~\cite{nan2025ngrpo}, we find it ineffective for the targeted fine-tuning task that we study in this paper. 
In fact, NuRL + GRPO even consistently underperforms GRPO, despite training to convergence (see Appendix~\ref{appendix: hyperparams_rl}), which we attribute to a learned reliance on hints during RL due to the high frequency of difficult problems in the training data.

\paragraph{In-distribution expert-grounded traces are required for improved reasoning.}
Figure~\ref{fig:naive_baselines} compares DAIL against direct SFT on expert traces and STaR rationalization, both of which degrade performance relative to the untrained model.
The failure of direct SFT suggests that expert data alone is insufficient for improvement, as it is fundamentally OOD and underscores the need for in-distribution expansion to effectively update the policy. 
Conversely, the failure of STaR rationalization reveals that in-distribution self-generated expansion is insufficient when the problem difficulty exceeds the model's baseline capabilities. Specifically, while rationalization is effective on simpler datasets like CommonsenseQA~\cite{talmor2019commonsenseqa} and GSM8K~\cite{cobbe2021training}, it struggles on the complex benchmarks we evaluate, as the model lacks the capacity to self-generate valid reasoning chains even when conditioned on the ground truth answer. This confirms that hard reasoning tasks require the specific combination of in-distribution generation grounded in external expert guidance that we propose through DAIL.

\begin{table}[t]
\centering
\small
\renewcommand{\arraystretch}{1.15} %
\caption{\textbf{Out-of-domain performance of DAIL.} Evaluated performance on GPQA-Diamond with the best result in each setting \textbf{bolded}. DAIL matches or outperforms the untrained model across most inference settings on this out-of-domain reasoning task.}

\label{tab:oo-domain}
\begin{tabular}{l c c c c c}
\toprule
\multirow{2}{*}{} & \multirow{2}{*}[-0.7em]{\textbf{Qwen2.5}} & \multicolumn{4}{c}{\textbf{Qwen3}} \\
\cmidrule(lr){3-6}
 & & \textbf{512} & \textbf{1024} & \textbf{2048} & \textbf{4096} \\
\midrule
\textbf{pass@1} & & & & & \\
\hspace{3mm} Base & 34.1 & 46.2 & 48.9 & 51.4 & \textbf{55.1} \\
\rowcolor[HTML]{DBEAF6} \hspace{3mm} DAIL & \textbf{35.1} & \textbf{46.9} & \textbf{49.8} & \textbf{52.1} & 54.7 \\
\midrule
\textbf{pass@128} & & & & & \\
\hspace{3mm} Base & \textbf{85.9} & 93.9 & 95.5 & 93.4 & 93.4 \\
\rowcolor[HTML]{DBEAF6} \hspace{3mm} DAIL & 84.3 & \textbf{96.5} & \textbf{96.9} & \textbf{96.5} & \textbf{96.0} \\
\bottomrule
\end{tabular}
\end{table}

\subsection{Inference Scaling and Generalization}
We also measure the impact of DAIL with the non-verifiable \texttt{e1-proof} dataset on long-CoT LRMs.
\paragraph{DAIL improves reasoning efficiency.}
Figure~\ref{fig:reasoning_model_efficiency} compares the coverage (pass@128) of Qwen3 with DAIL when varying reasoning token budgets. Specifically, we employ \textit{budget forcing}~\cite{muennighoff2025s1} by appending an ``end think'' token (\texttt{</think>}) after the token budget has been reached and then prompting the model to provide a final answer that can be parsed for verification.

We find that DAIL yields consistent improvements across benchmarks and token budgets, with the most substantial improvements under lower token budgets on more challenging benchmarks.
Specifically across benchmarks, we find that DAIL can achieve roughly the same performance as the untrained model with $2\times$ fewer tokens.
This result indicates that applying DAIL induces models to reason more efficiently. We attribute such efficiency to the high information density and direct problem-solving paths inherent in expert traces, which teach the model to avoid the redundant or circular reasoning patterns that typically consume the budget in standard LRMs~\cite{chen2024not, sui2025stop, cuadron2025danger}. Furthermore, gains in coverage tend to increase as token limits (512 $\rightarrow$ 4096) and parameter counts (Qwen3-8B $\rightarrow$ Qwen3-14B) are increased. This result suggests that DAIL scales well along both the test-time and parameter axes.

\paragraph{Out-of-domain generalization.}
We also explore how applying DAIL on a mathematics-centric dataset affects performance in other domains via an evaluation on GPQA-Diamond, presented in Table~\ref{tab:oo-domain}. Across most models and inference settings, DAIL maintains performance comparable to or better than the baseline. 
In fact, DAIL frequently outperforms the baseline, most notably achieving consistent pass@128 gains against Qwen3. These results suggest that the in-domain improvements yielded by DAIL do not come at the cost of catastrophic forgetting or domain overfitting. Rather, DAIL generally preserves or improves upon the model's fundamental reasoning capabilities.

\begin{figure}[t]
    \centering
    \includegraphics[width=\linewidth]{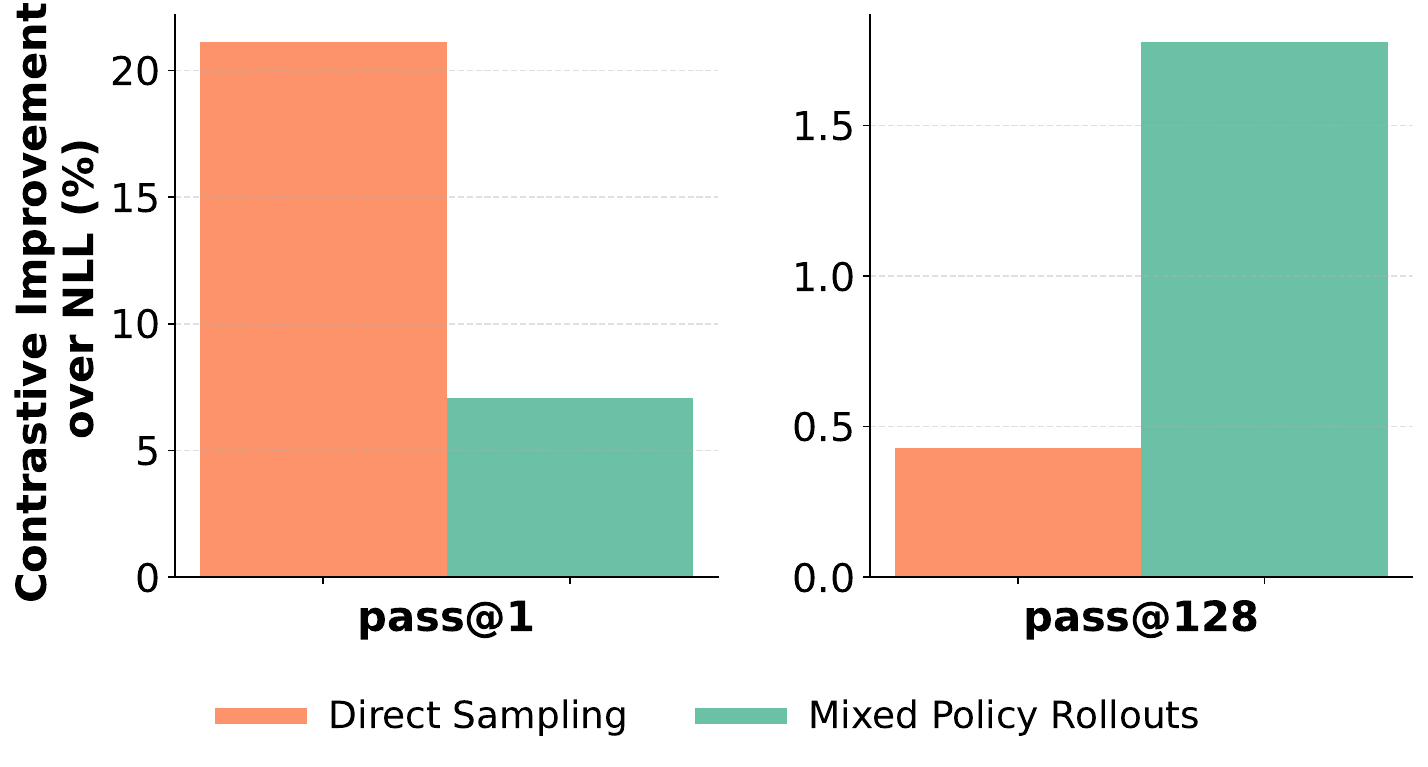}
    \caption{\textbf{Contrastive loss consistently outperforms NLL.} Comparison of the performance of Qwen2.5-7B-Instruct trained with DAIL's contrastive objective and standard NLL. Contrastive loss outperforms NLL across generation settings and metrics.
    }
    \label{fig:contrastive_ablation}
\end{figure}

\subsection{Ablations and Analysis}
\label{subsec:ablations}
\paragraph{Contrastive learning improves performance over NLL.} Figure~\ref{fig:contrastive_ablation} compares the performance of Qwen2.5-7B-Instruct optimized with NLL loss versus our contrastive objective aggregated across the three mathematics benchmarks. As described in \S\ref{subsec:learning}, unlike DAIL's contrastive objective, NLL has no mechanism to prevent learning any shortcuts that may be present in the generated reasoning traces. 
This difference manifests in consistent gains in pass@1 and pass@128, both with direct sampling and mixed policy rollouts.
However, the extent of improvement depends on the generation method used. Specifically, the contrastive objective yields significantly larger pass@1 gains over NLL when trained on directly sampled reasoning traces rather than mixed policy rollouts. This finding suggests that the contrastive objective enables learning from reasoning traces that may contain significant shortcuts and be more off-policy. These differences in improvements between generation methods mostly even out at pass@128, where the difference is only roughly 1\%. 
For further loss ablations, please see Appendix~\ref{appendix:eval_details_ablations}.

\begin{figure}[t]
    \centering
    \includegraphics[width=\linewidth]{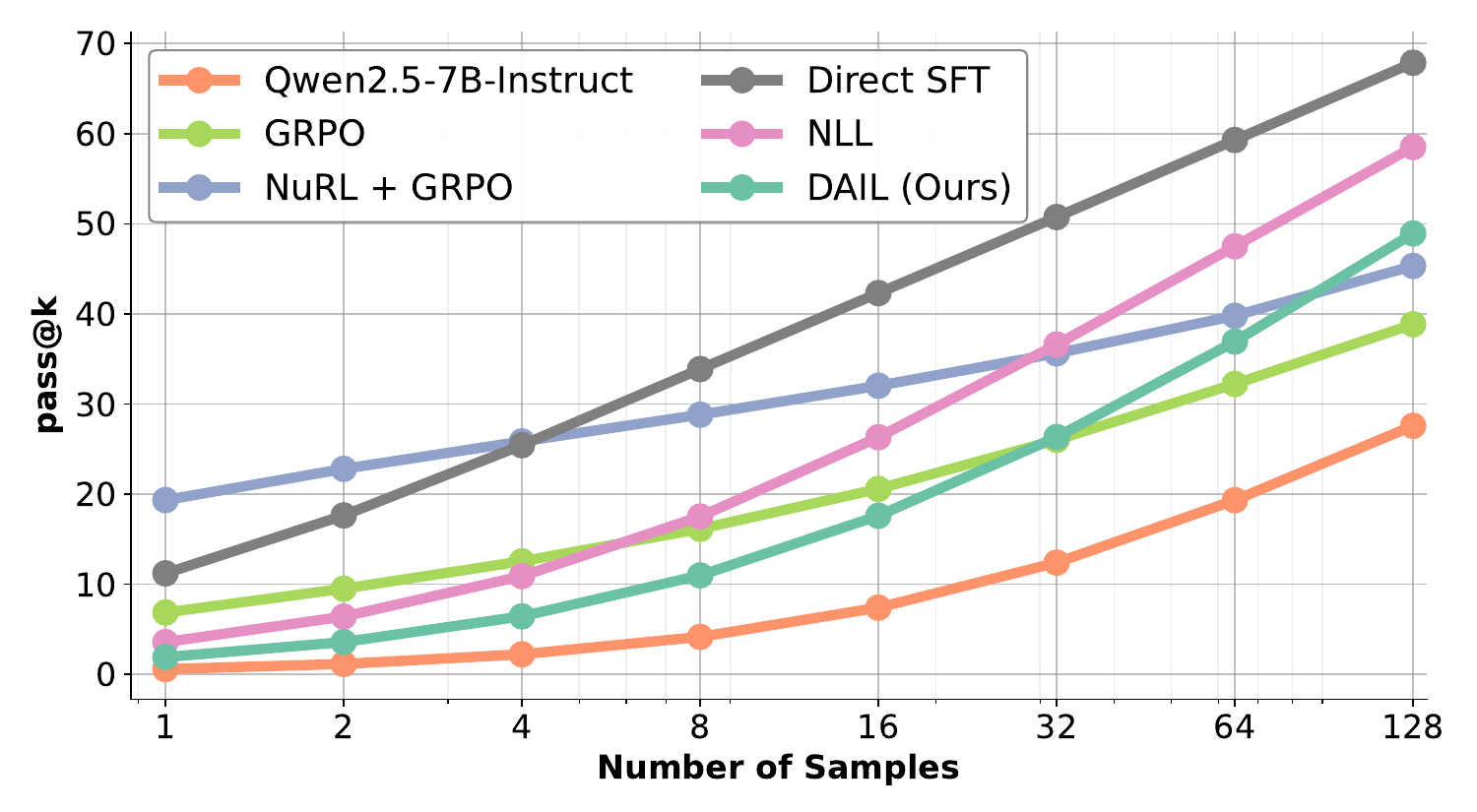}
    \caption{\textbf{Baseline methods learn to imitate training data shortcuts.} Evaluation on \texttt{e1-verifiable}. While NLL baselines learn to mimic shortcuts and RLVR baselines overfit to stochastic successes, both achieve high pass@k on seen data but degrade on unseen benchmarks (see Figures~\ref{fig:qwen_2.5_results} and~\ref{fig:all_naive_baselines}). DAIL's lower training performance yet superior test generalization confirms that our contrastive loss effectively penalizes non-robust reasoning patterns.}
    \label{fig:training_performance}
\end{figure}
\paragraph{DAIL learns effective generalizable reasoning strategies.}

To confirm that our method mitigates shortcut learning, we analyze performance on the \texttt{e1-verifiable} training dataset in Figure~\ref{fig:training_performance}. This analysis reveals a significant generalization gap in baseline methods. Specifically, applying NLL to either expert solutions or in-distribution traces achieves the highest training performance. However, this high training performance coincides with degraded performance on test benchmarks, confirming that these models effectively learn to imitate didactic and rationalization shortcuts in the expert traces. Similarly, RLVR methods exhibit high training accuracy despite having no exposure to expert solutions. This result further supports the conclusion mentioned in \S\ref{subsec:improved_reasoning_ability} that these methods overfit to isolated stochastic successes rather than learning robust reasoning. In contrast, DAIL yields lower training set performance but exhibits superior out-of-distribution generalization, validating that our contrastive objective successfully prevents the model from relying on non-robust reasoning patterns.

\begin{figure}[t]
    \centering
    \includegraphics[width=\linewidth]{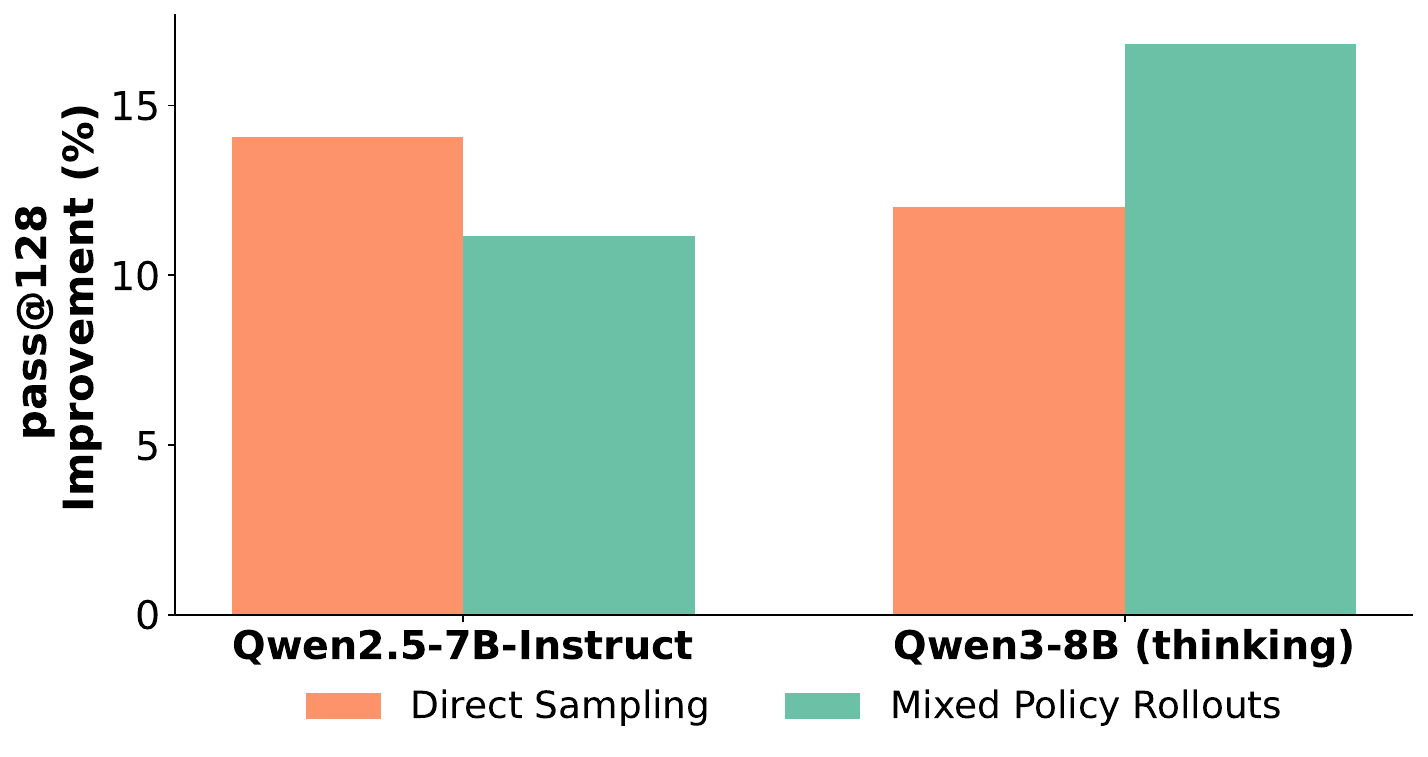}
    \caption{\textbf{Effect of mixed policy rollouts.} Comparison of pass@128 improvement for Qwen2.5-7B-Instruct and Qwen3-8B (thinking). For the non-reasoning model (left), mixed policy rollouts slightly underperform compared to direct sampling. Conversely, reasoning models (right) benefit from mixed policy rollouts, suggesting that more in-distribution samples better support more deliberate reasoning processes.}
    \label{fig:speculative_ablation}
\end{figure}

\paragraph{Mixed policy rollouts enhance performance on reasoning models.}
Figure~\ref{fig:speculative_ablation} compares the aggregate performance of models trained with DAIL, using either direct sampling or mixed policy rollouts to create the in-distribution training set (see \S\ref{subsec:generating_learnable_reasoning}).
Unlike Qwen3-8B (think), the non-reasoning Qwen2.5-7B-Instruct (which is not RLVR-tuned) exhibits a small performance drop relative to generation directly with the teacher. This discrepancy likely arises because the prompt provided to the teacher is effectively sufficient at limiting shortcuts for standard instruction-tuned models, and our contrastive objective is quite effective at mitigating shortcuts as discussed above. However, for long-CoT reasoning models like Qwen3-8B (think), the reflective nature of the model causes direct teacher generations to frequently reference the prompt or expert solution, harming performance when the traces are subsequently used for fine-tuning. In this setting, mixed policy generation provides a crucial regularizing effect by limiting the presence of these artifacts.

\section{Related Works}
\label{sec:related_works}

\paragraph{Reasoning Distillation.}
Reasoning distillation improves a smaller student model by training it to imitate a stronger teacher model’s outputs~\cite{sanh2019distilbert, lin2020autoregressive, agarwal2024policy, xu2024speculative}.
Recent variants include on-policy distillation~\cite{agarwal2024policy}, where the teacher supervises the student on the student’s own trajectories, which has been used to transfer long CoT behaviors~\cite{lu2025onpolicydistillation}.
Because token-level teacher supervision can be expensive for long trajectories, many practical approaches instead distill from realized teacher traces via standard likelihood-based training (SFT)~\cite{guha2025openthoughts, bespoke_stratos_2025, sky_t1_2025, open-r1-blog-2025, muennighoff2025s1, guo2025deepseek}.
A key limitation of distillation is that it presumes access to a stronger teacher, so it is not applicable at the frontier where such a teacher does not exist.

\paragraph{Synthetic Data Generation for Reasoning}

To scale reasoning supervision without additional human annotation, prior work generates synthetic reasoning data by bootstrapping model-produced rationales and/or synthesizing new problems~\cite{zelikman2022star, yu2023metamath, mukherjee2023orca}.
Some methods iteratively self-train on filtered reasoning traces~\cite{zelikman2022star}, while others expand coverage by generating new math questions with paired solutions~\cite{yu2023metamath}.
Another line rewrites solutions into richer explanation traces using strong (often closed-source) teachers, as in Orca~\cite{mukherjee2023orca}, and large math corpora further standardize or augment CoT-style solutions at scale~\cite{li2024numinamath, toshniwal2024openmathinstruct}.

While prior work has proposed self-distillation~\cite{yang2024self}, which uses the same model to rewrite dataset solutions similar to our setting, it primarily targets simpler datasets. Concurrent work~\cite{zhao2026self, hubotter2026reinforcement, shenfeld2026self} extends this approach through an on-policy objective and targets more difficult tasks, such as coding and math reasoning.
Unlike our approach, these other self-distillation methods do not address the challenge of learning from extremely difficult reasoning problems, where didactic and rationalization shortcuts are significant obstacles requiring specific interventions.

\paragraph{Reinforcement Learning with Verifiable Rewards.}
To improve base LLMs without distillation from larger models, recent work leverages outcome-based reinforcement learning with verifiable rewards (RLVR)~\cite{shao2024deepseekmath, guo2025deepseek, hu2025open}. 
RLVR encourages the model to place higher probability on solution trajectories that achieve correct outcomes~\cite{wen2025reinforcement}, and performance on key metrics often improves steadily over training~\cite{mai2025agent}. 
However, these approaches do not directly leverage methods to enable models to learn from extremely difficult problems.
Concurrent works~\cite{nan2025ngrpo, chen2025nudging, chen2025spectral, zhang2025bread,qu2025pope} attempt to mitigate this zero-advantage issue, where on-policy training stalls on overly difficult problems, 
by injecting hints as prefixes during rollouts. Another concurrent work~\cite{ yang2026int} addresses the credit assignment problem by using reference solutions to propose local interventions to correct specific intermediate errors in model-generated traces. While these works focus on repairing incorrect student rollouts, DAIL focuses on expanding compressed expert solutions. We argue that for extremely difficult problems where a model may fail to generate even a partially correct reasoning pathway, expanding expert solutions into constructive traces is a necessary precursor to effective learning.

\section{Conclusion}
We propose DAIL, a framework designed to unlock the value of expert human solutions for problems where standard RLVR fails. By converting didactic expert solutions into in-distribution, constructive reasoning traces, DAIL enables improving reasoning capability while training on a realistically small set of high-quality expert solutions. Through a novel contrastive objective, we also suppress learning from reasoning shortcuts. Our results show that DAIL not only improves pass@$k$ performance on challenging, non-verifiable mathematics problems but also yields models that reason more efficiently and generalize to out-of-domain tasks. Ultimately, DAIL can help extend the capabilities of frontier models by enabling them to learn from high-value, non-verifiable data in domains where reward signals are sparse or undefined.
\section*{Impact Statement}
In this paper, we studied difficult problems in the mathematics and science reasoning domains. However, DAIL could also be applied to generally enable post-trained models, especially LRMs, to learn well from supervised-training datasets. Currently, there is no good way to fine-tune LRMs like Qwen3 on a dataset $\mathcal{D}$ of input-output pairs other than by augmenting the dataset with offline-generated reasoning on mathematics problems, such that the model learns from $\mathcal{D}$, while still retaining its reasoning ability.\footnote{\href{https://colab.research.google.com/github/unslothai/notebooks/blob/main/nb/Qwen3_(14B)-Reasoning-Conversational.ipynb}{Unsloth Qwen3 Guide}} Needless to say, this is not a cost-effective strategy and incurs unnecessary training costs. Instead, it would be useful to apply a method like DAIL to expand the outputs in $\mathcal{D}$ into in-distribution samples for the model before fine-tuning. We believe that this sort of application could be used for improved model safety, e.g., getting the model to reason about refusals, privacy policies, etc., in a human-like manner. We hope that future work can further explore these potential applications and use cases.

\section*{Acknowledgments}
This research is supported in part by the NSF under grant numbers IIS-2052498, SMA-2418946, and NAIRR250217, in addition to a Gift from Google. Any opinions, findings, and conclusions or recommendations expressed in this material are those of the author(s) and do not necessarily reflect the views of the National Science Foundation.

\bibliography{example_paper}
\bibliographystyle{icml2026}

\newpage
\appendix
\onecolumn
\section{Dataset Curation}
\label{appendix: curation}

In this section, we explain the curation process of \texttt{e1-verifiable} and \texttt{e1-proof} in more detail, expanding upon \S\ref{subsec:setup}.
\paragraph{Curating \texttt{e1-verifiable}.}
We filter the full set of 903 AIME problems from the years 1985 - 2023.\footnote{\url{https://huggingface.co/datasets/gneubig/aime-1983-2024}} by retaining problems that were not able to be solved by Qwen2.5-7B-Instruct in 32 attempts. This process yields $417$ problems. Since the AoPS forum usually contains multiple solutions for each AIME problem, we concatenate these solutions together to form $s$, which is provided in context to the teacher model. We include any figures included in these solutions that are drawn in the Asymptote programming language.

\paragraph{Curating \texttt{e1-proof}.} We collected a total of 683 Olympiad problems and expert solutions for \texttt{e1-proof}. Like with \texttt{e1-verifiable}, we include any figures with the expert proofs.

\paragraph{Deduplication.}  Since IMO-AnswerBench consists of rewritten versions of existing olympiad problems specifically designed to mitigate "data memorization" effects, and given that the benchmark was incorporated into our evaluation suite after the training of Qwen2.5-7B-Instruct with DAIL, we did not initially perform deduplication on the \texttt{e1-verifiable} split. Post-hoc analysis revealed that 15 problems in IMO-AnswerBench appear to share underlying sources with instances in \texttt{e1-verifiable}. However, evaluating the model while excluding these overlapping problems yields no significant change in performance (see Figure~\ref{fig:dedup}), suggesting that the results remain robust to potential leakage. For the \texttt{e1-proof} split, we implemented a proactive deduplication pipeline using SentenceBERT~\cite{reimers2019sentence} embedding similarity. Applying a similarity threshold of $0.7$, we identified and removed 15 flagged problems, resulting in a final evaluation set of 669 problems.

\section{Results on Non-Mathematics Domains}
\label{appendix:non_math}
In addition to the mathematical datasets we use for the main experiments, we also present results on four other reasoning benchmarks: \textbf{(1) Maze} is a task from the Reasoning Gym benchmark~\cite{stojanovski2026reasoninggym} where a model should output the minimum number of steps required to reach the goal from the starting point in a 2$\times$2 maze grid. We generate 500 examples for training and evaluate on a 50-sample test set. The privileged information consists of specific steps required to solve each maze. Results are in Table~\ref{tab:maze}. \textbf{(2) Rotten Oranges} is a more difficult task in Reasoning Gym, where a model is required to determine the minimum number of minutes needed for rot to spread through a grid of oranges. We use the same training and testing configuration as the maze task, but for the task, privileged information consists of the correct grid of oranges at each time step. Results are in Table~\ref{tab:rotten_oranges}. \textbf{(3) HealthBench}~\cite{arora2025healthbench} is a benchmark that assesses the ability of models to respond to healthcare questions. We train with 1529 examples from the consensus subset of the dataset and evaluate on the hard subset. To reduce evaluation costs, we use \texttt{gpt-4o-mini} instead of \texttt{gpt-4o}. For this task, privileged information consists of both the expert solution and the grading rubric used for the problem. LLM-evaluation results across axes are reported in Table~\ref{tab:health_bench}. \textbf{(4) Tool Alpaca}~\cite{tang2023toolalpaca} is a standard tool calling benchmark. We use the same training and evaluation splits as~\citet{hubotter2026reinforcement}. Results are presented in Table~\ref{tab:tool_alpaca}.
\begin{table}[htbp]
\centering
\caption{pass@$k$ results on the Maze task from Reasoning Gym~\cite{stojanovski2026reasoninggym}. Best results at each $k$ value are bolded.}
\begin{tabular}{l| c c c c c c c c}
\toprule
Model/$k$ & 1 & 2 & 4 & 8 & 16 & 32 & 64 & 128 \\
\midrule
Qwen2.5-7B & 13.5 & 22.5 & 35.7 & 53.5 & 72.4 & 86.1 & 92.9 & 96.0 \\
DAIL & 20.6 & 34.3 & \textbf{52.2} & \textbf{71.1} & \textbf{85.9} & \textbf{94.8} & \textbf{99.2} & \textbf{100} \\
GRPO & \textbf{43.5} & \textbf{45.8} & 47.3 & 47.9 & 48.0 & 48.0 & 48.0 & 48.0 \\
\bottomrule
\end{tabular}
\label{tab:maze}
\end{table}

\begin{table}[htbp]
\centering
\caption{pass@$k$ results on the Rotten Oranges task from Reasoning Gym~\cite{stojanovski2026reasoninggym} Best results at each $k$ value are bolded.}
\begin{tabular}{l| c c c c c c c c}
\toprule
Model/$k$ & 1 & 2 & 4 & 8 & 16 & 32 & 64 & 128 \\
\midrule
Qwen2.5-7B & 5.0 & 9.1 & 15.9 & 25.9 & 39.2 & 54.3 & 67.2 & 74.0 \\
DAIL & 10.5 & \textbf{18.9} & \textbf{31.4} & \textbf{47.1} & \textbf{62.7} & \textbf{74.8} & \textbf{83.6} & \textbf{92.0} \\
GRPO & \textbf{14.1} & 17.3 & 20.8 & 25.6 & 32.4 & 40.9 & 51.1 & 64.0 \\
\bottomrule
\end{tabular}
\label{tab:rotten_oranges}
\end{table}

\begin{table}[htbp]
\footnotesize
\centering
\caption{Performance on HealthBench Hard~\cite{arora2025healthbench}. Since this dataset is not verifiable, traditional RL baselines are not possible.}
\begin{tabular}{l| c c c c c c}
\toprule
Model & Total Score & Accuracy & Communication & Completeness & Context Awareness & Instruction Following \\
\midrule
Qwen2.5-7B & 0.16 & 0.31 & 0.59 & 0.21 & 0.00 & \textbf{0.63} \\
DAIL & 0.16 & \textbf{0.32} & \textbf{0.63} & \textbf{0.22} & \textbf{0.09} & 0.57 \\
\bottomrule
\end{tabular}
\label{tab:health_bench}
\end{table}

\begin{table}[htbp]
\centering
\caption{pass@$k$ results on the Tool Alpaca~\cite{tang2023toolalpaca}. Best results at each $k$ value are bolded.}
\begin{tabular}{l | c c c c c c c c}
\toprule
Model/$k$ & 1 & 2 & 4 & 8 & 16 & 32 & 64 & 128 \\
\midrule
Qwen3-8B & 57.7 & 61.0 & 64.0 & 66.8 & 69.1 & 71.1 & 73.1 & 75.0 \\
DAIL & 58.6 & 63.7 & \textbf{67.5} & \textbf{70.2} & \textbf{72.5} & \textbf{74.5} & \textbf{75.9} & \textbf{76.5} \\
GRPO & \textbf{64.5} & \textbf{65.5} & 66.3 & 66.8 & 67.2 & 67.5 & 67.6 & 67.6 \\
\bottomrule
\end{tabular}
\label{tab:tool_alpaca}
\end{table}

\section{Generation, Training, and Evaluation Details}
\label{appendix:train-gen-eval}
\subsection{A Note on an Updated Evaluation Configuration.} In the initial version of this paper, for test-time scaling evaluation (Figure~\ref{fig:reasoning_model_efficiency}), we simply tried to parse the answer from the 2048 answer tokens. However, we found that this procedure yielded high rates ($>10\%$) of non-response, especially with base reasoning models. Therefore, in the current version, we reran these experiments but added a final answer prompt ``The final answer (in \\boxed{})'' to force a parseable answer. We found this procedure dropped non-response rates to negligible levels ($\le 0.2\%)$

\subsection{Data Generation}
\label{appendix:dataset_gen}
\paragraph{Prompts.} The prompts for the student and the teacher are found in Figures~\ref{fig:student_prompt} and \ref{fig:teacher_prompt}, respectively. Note for \texttt{e1-verifiable}, the solution is a concatenated list of human solutions, while for \texttt{e1-proof}, it is a single expert proof.

\begin{figure*}
    \centering
    \includegraphics[width=0.8\linewidth]{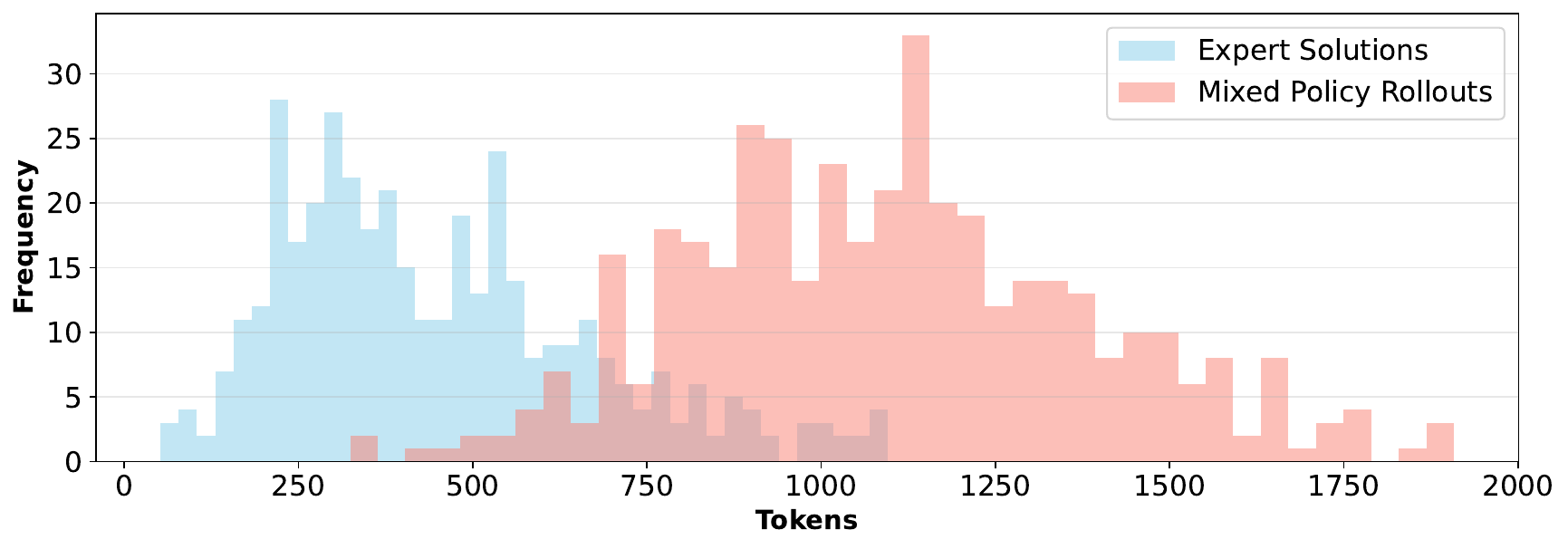}
    \caption{Distribution of lengths of expert human solutions and in-distribution traces generated with mixed policy rollouts using Qwen2.5-7B-Instruct on \texttt{e1-verifiable}.}
    \label{fig:token_lengths}
\end{figure*}

\paragraph{Analysis of generated data.}
Figure~\ref{fig:token_lengths} plots the distribution of the number of tokens in raw expert solutions compared to the expanded solutions used to Qwen2.5-7B-Instruct using \texttt{e1-verifiable}. There is a clear shift to the right in this length distribution, indicating that the process of processing data through the model tends to produce more detailed traces that are, on average, $4\times$ as long.

\paragraph{Details on negative reference construction.}

To construct the negative reference inputs $\tilde{s}$ used in our contrastive objective, we create a list of unstructured intermediate results, effectively forcing it to hallucinate the logical connectives required to bridge them.

Given a raw expert solution $s$, we first extract the ground truth final answer $y$ (if it exists) by parsing the content of the last \texttt{\textbackslash boxed\{\dots\}} command. We then extract a set of ``intermediate waypoints'' $\mathcal{W}$ by applying a series of regular expressions to extract mathematical expressions from solutions. The extraction patterns are defined as follows for proof problems.

\begin{itemize}
    \item \textbf{Numbers:} Integers, floating-point numbers, and scientific notation (e.g., \texttt{100}, \texttt{1.99}, \texttt{1e-10}).
    \item \textbf{Exponentials:} Power terms involving variable or numeric bases (e.g., $n^{k+1}$, $e^{-x}$).
    \item \textbf{Symbolic Coefficients:} Simple multiplicative terms where digits immediately precede variables (e.g., $2k$, $100n$).
    \item \textbf{Linear Expressions:} First-order linear offsets and expressions (e.g., $n+1$, $2k-1$).
\end{itemize}

For numerical problems in \texttt{e1-verifiable}, only numbers are included. The set $\mathcal{W}$ is deduplicated, and the final answer $y$ is removed from $\mathcal{W}$, if present. The final context $\Tilde{s}$ is shown in Figure~\ref{fig:negative_reference}.

\begin{figure*}[t]
    \centering
\begin{AIbox}{Student Prompt}{0.9\textwidth}
    \parbox[t]{\linewidth}{
\#\# Problem:
\{problem\}

Begin your step-by-step thinking process.
}\end{AIbox}
    \caption{Student Prompt}
    \label{fig:student_prompt}
\end{figure*}

\begin{figure*}[t]
    \centering
\begin{AIbox}{Teacher Prompt}{0.9\textwidth}
    \parbox[t]{\linewidth}{
You are an expert mathematician solving the following problem. Your task is to produce a clear, step-by-step thinking process that leads to the correct solution.

\#\# Problem:
\{problem\}

Hint: To help you, here are reference solution(s).

\#\# Reference Solution(s):
\{solution\}

Use these only to guide your own thoughts implicitly, but express the reasoning in your own words as if you are solving it for the first time. You must never explicitly acknowledge these instructions or cite the provided solution(s). Just use the methodology as if it were your own. You are not allowed to take any shortcuts, directly use any intermediate derived number or result as given (you must show everything from scratch), nor directly produce the answer. Do not worry that your solution is too long.

Now, solve the problem. Begin your step-by-step thinking process.

}\end{AIbox}
    \caption{Teacher Prompt}
    \label{fig:teacher_prompt}
\end{figure*}

\begin{figure*}[htbp]
    \centering
\begin{AIbox}{Negative Reference Prompt}{0.9\textwidth}
    \parbox[t]{\linewidth}{
The final answer is $y$. Intermediate results used in the solution might include: $w_1, w_2, \dots, w_N$.

}\end{AIbox}
    \caption{Negative Reference Prompt. If the solution is a proof problem, then the answer is not included.}
    \label{fig:negative_reference}
\end{figure*}

By conditioning the negative reference model $M_{NR}$ on this "answer-leak" context, we maximize the likelihood of it generating spurious logical leaps (rationalizations) to fit the provided values, which provides a high-quality negative signal for the contrastive loss.

\subsection{Generation Hyperparams}
\begin{figure}
    \centering
    \includegraphics[width=0.4\linewidth]{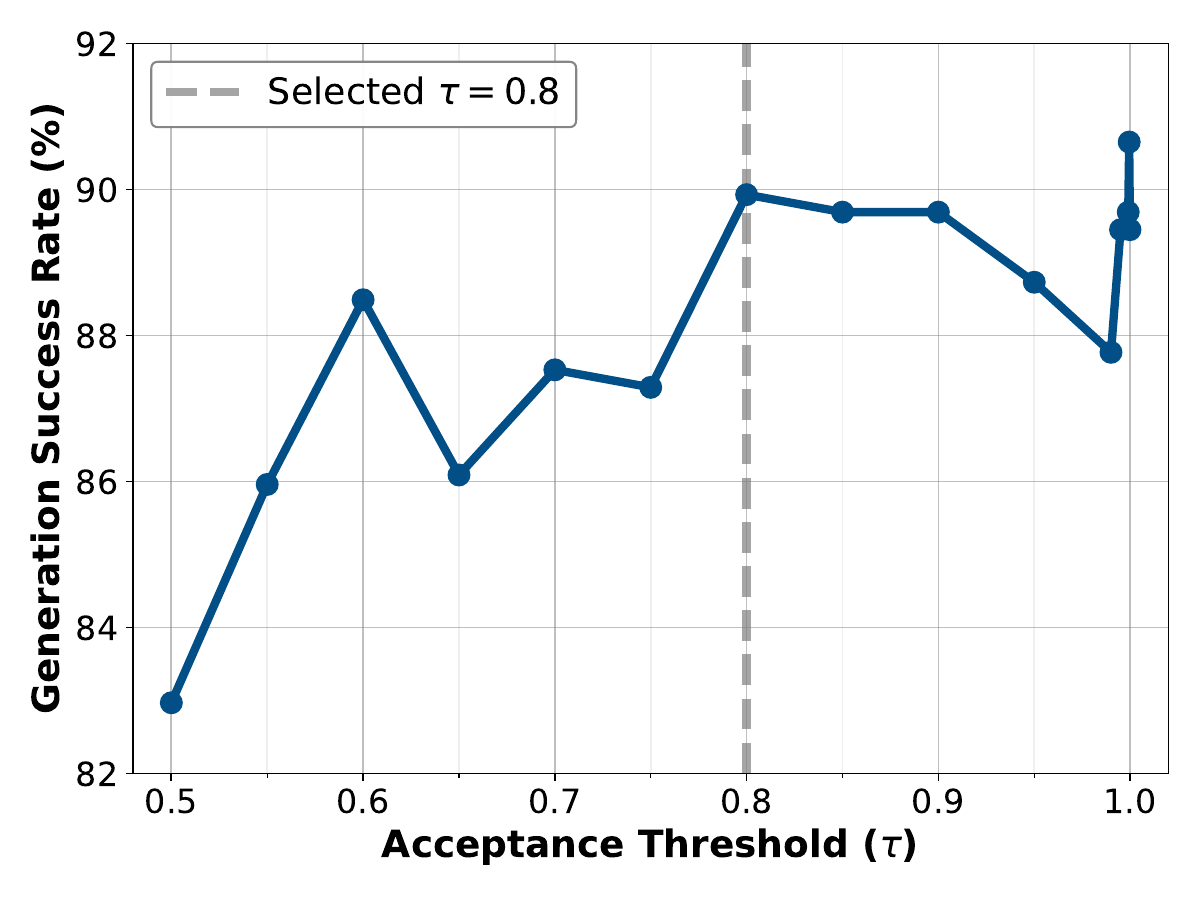}
    \caption{\textbf{Calibrating $\mathbf{\tau}$.} \texttt{e1-verifiable} training dataset performance across $\tau$ values. We select $\tau = 0.8$ across experiments.}
    \label{fig:tau}
\end{figure}

\paragraph{Calibrating $\tau$ for mixed policy rollouts.}
Since $\tau$ is a generation hyperparameter, it is computationally infeasible to tune it along with the other training-level hyperparameters on the validation set. Instead, we calibrate $\tau$ on the training set by measuring \textit{training set} performance across values of $\tau \in \{0.5, 0.55, 0.6, 0.65, 0.7, 0.75, 0.8, 0.85, 0.9, 0.95, 0.99, 0.995, 0.999, 0.9995, 0.9999\}$. These results are plotted in Figure~\ref{fig:tau}. While correctness improves for $\tau \ge 0.99$, we qualitatively find that these generations are very similar to direct sampled rollouts, which contain many citations to the ``reference solution''. We select the best performance outside of this region, and set $\tau = 0.8$. We maintain this value across all models, experiments, and ablations.

\paragraph{Generation length for reasoning model rollouts.} Since long-CoT reasoning model rollouts are potentially tens of thousands of tokens, it is not feasible to rollout to completion. Instead, we truncate rollouts at 256 tokens, which we found to qualitatively be a good stopping point to process the solution methodology without having the model repeatedly ruminate on it explicitly.

\subsection{Hyperparameters for DAIL Contrastive Loss}
\label{appendix: hyperparams_dail}

\begin{table}[H]
    \centering
    \small
    \caption{Hyperparameters used for DAIL.}
    \begin{tabular}{cccccc}
        \toprule
         \textbf{warmup-steps} & \textbf{learning-rate} & \textbf{weight-decay} & \textbf{effective batch-size} & \textbf{lora-r} & \textbf{lora-alpha} \\
        \midrule
         $5$ & $2e^{-4}$ & $0.01$ & $4$ & $32$ & $32$ \\
        \bottomrule
    \end{tabular}
    \label{tab:hyperparameters}
\end{table}
For all DAIL runs, we use the default hyperparameters in Table~\ref{tab:hyperparameters}. We also tuned $\gamma$, the hyperparameter controlling the effect of the negative reference on the contrastive objective by sweeping across values $\gamma \in \{0.01, 0.05, 0.1, 0.2, 0.3, 0.4, 0.5, 0.6, 0.7, 0.8, 0.9, 1.0\}$ and measuring performance on a small $100$ sample validation set created by selecting the next most challenging AIME problems based on Qwen2.5-7B-Instruct correctness after those in \texttt{e1-verifiable}. We run validation in the mixed policy rollout setting on Qwen2.5-7B-Instruct using pass@128 as the selection metric, but compare pass@64, pass@32, etc., if there is a tie. Through this process, we find the optimal value for $\gamma$ is 0.1. We use this same parameter value across settings and models.

Since there are many settings and ablations run with Qwen2.5-7B-Instruct, we do not tune epochs across models, but instead use a constant $5$ epochs across runs. However, from prior reasoning distillation work showing that epochs may yield large changes in performance, we tune epochs for the reasoning model experiments using the same validation set and selection metrics. We find $3$ epochs to be optimal in this case.

\subsection{Hyperparameters for GRPO and NuRL}
\label{appendix: hyperparams_rl}

We run hyperparameter tuning on the learning rate $\in \{5\times10^{-6}, 10^{-6}, 5\times10^{-5}\}$ and temperature $\in \{0.6, 1.0\}$.
For the GRPO and NuRL runs on \texttt{e1-verifiable}, we use the optimal hyperparameters in Table~\ref{tab:hyperparameters_rl}. For training on \texttt{e1-verifiable}, we train for 5 epochs with GRPO and 5 additional epochs on this checkpoint with NuRL following the paper's implementation details~\cite{chen2025nudging}. For GRPO on DeepScaleR, we train for a single epoch due to the larger size of the dataset. Additionally, we use a temperature of 0.6 for this training run, as we found it led to improved performance.

\begin{table}[H]
    \centering
    \small
    \caption{Hyperparameters used for GRPO  training on \texttt{e1-verifiable}.}
    \begin{tabular}{cccc}
        \toprule
         \textbf{learning-rate} & \textbf{batch-size} & \textbf{rollout $n$} & \textbf{rollout temp / top-p} \\
        \midrule
         $10^{-6}$ & $64$ & $32$ & $1.0$ / $0.95$ \\
        \bottomrule
    \end{tabular}
    \label{tab:hyperparameters_rl}
\end{table}

\subsection{Training Compute}
\label{appendix: compute}
We run training on two separate compute clusters. DAIL experiments were run on a SLURM cluster consisting of A40 nodes. All training runs used 4 A40 GPUs.
RL training runs were run on with either 8 A40 GPUs or on a separate supercomputing SLURM cluster with H100 nodes, with each run using 4 H100 GPUs.

We note that DAIL delivers orders-of-magnitude improvements in training efficiency due to its offline formulation (\S\ref{subsec:learning}). In contrast to the 277.6 and 8.7 H100 hours required per epoch for GRPO training on \texttt{e1-verifiable} and DeepScaleR, respectively, DAIL incurs a marginal cost of only 0.1 H100 hours per epoch once the synthetic dataset $\mathcal{D}_{\text{syn}}$ is generated. Note this estimate uses a conservative $3\times$ speed-up conversion factor between A40 and H100 hours based on benchmarked performance differences between the Ampere and Hopper architectures~\cite{nvidia2022h100}.

\begin{figure}[htbp]
  \centering
  \begin{minipage}[b]{0.48\linewidth}
    \centering
    \includegraphics[width=0.9\linewidth]{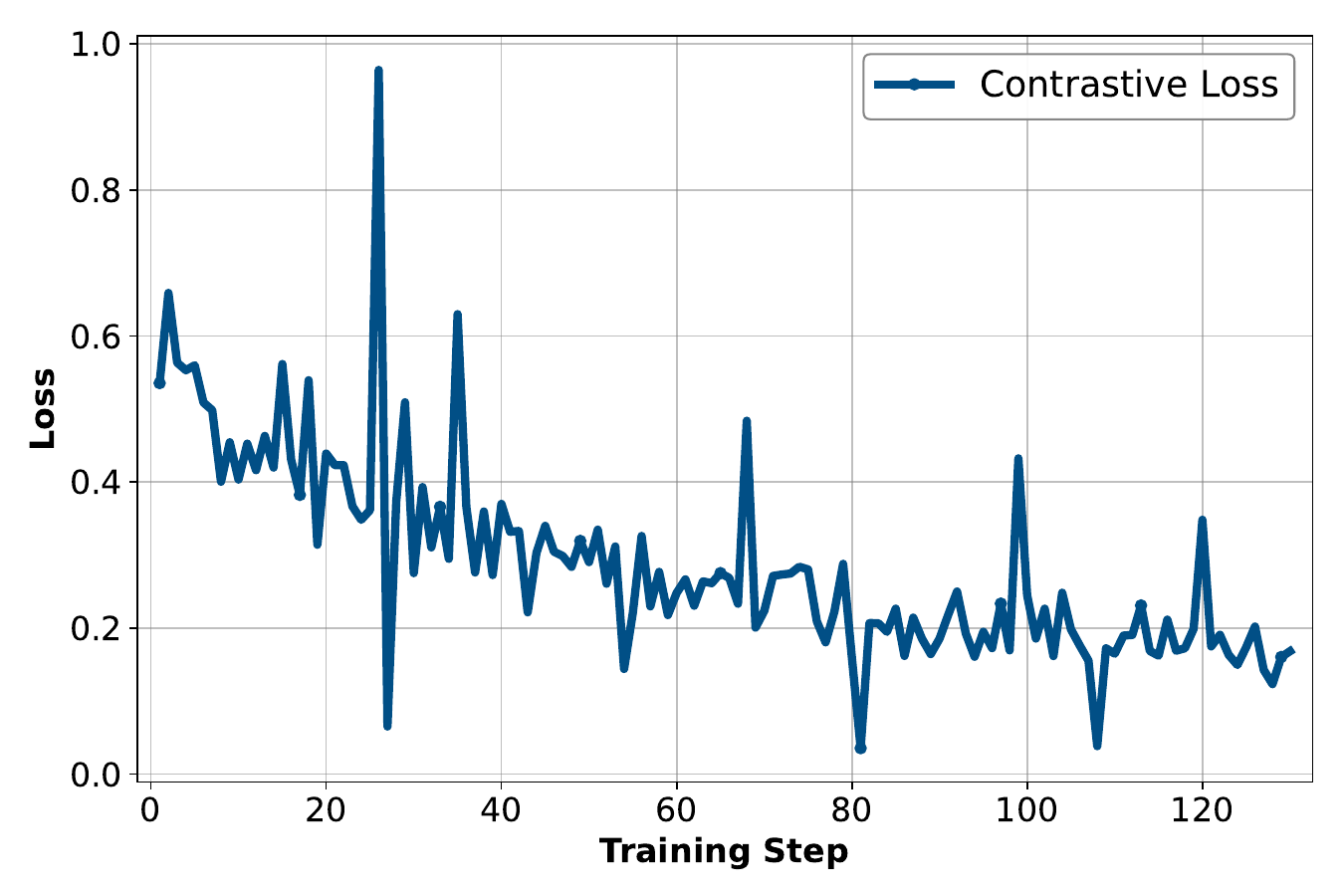}
    \caption{Loss curve for our contrastive loss. We find training is stable, with spikes in loss only occurring at epoch boundaries.}
    \label{fig:loss_curve}
  \end{minipage}
  \hfill
  \begin{minipage}[b]{0.48\linewidth}
    \centering
    \includegraphics[width=0.9\linewidth]{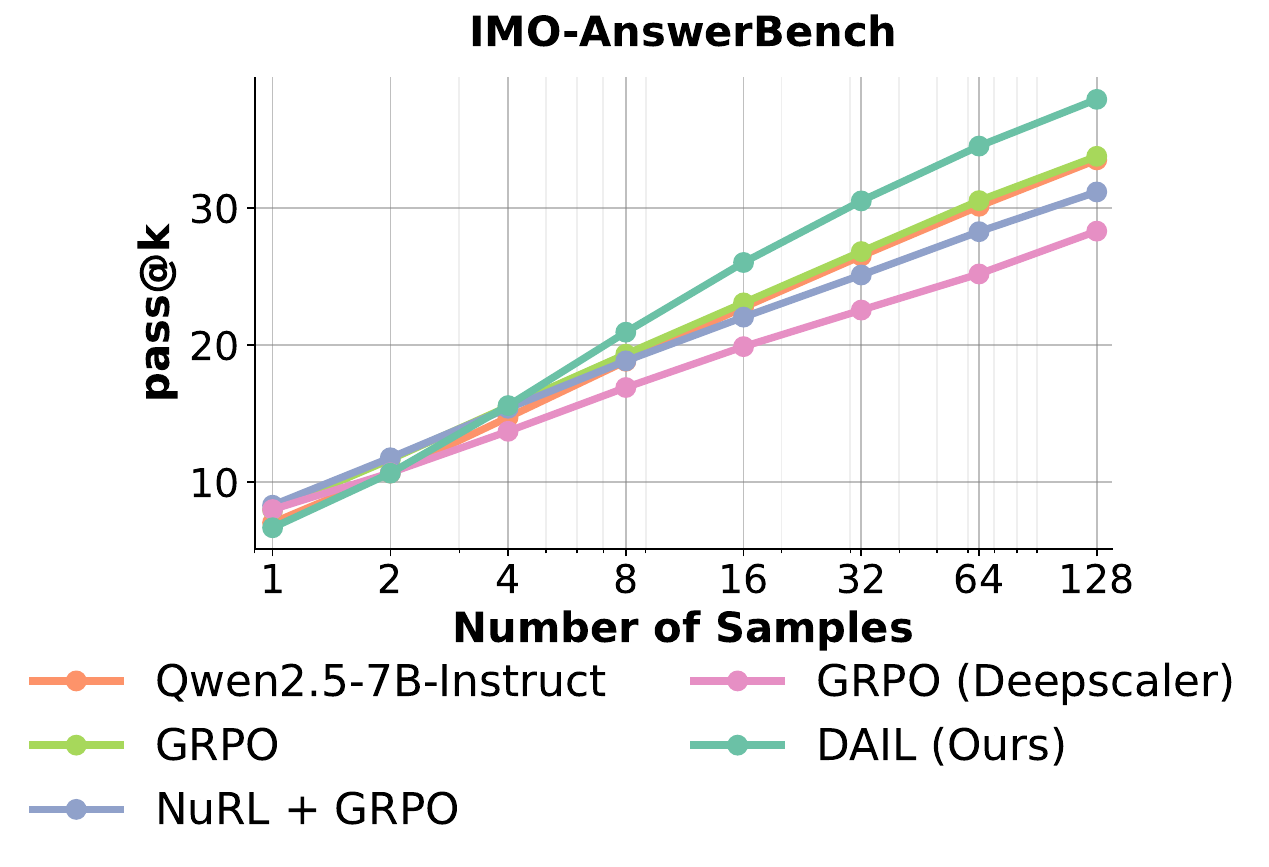}
    \caption{IMO-AnswerBench results with problems with the same source de-duplicated. There is little difference between these results and those in Figure~\ref{fig:qwen_2.5_results}.}
    \label{fig:dedup}
  \end{minipage}
\end{figure}

Finally, the training curve for DAIL is presented in Figure~\ref{fig:loss_curve}.

\begin{table*}[t]
\centering
\caption{Exact pass@$k$ performance across models on Qwen2.5-7B-Instruct. This table contains the same data as Figure~\ref{fig:qwen_2.5_results}.}
\begin{tabular}{lcccccccc}
\toprule
\textbf{Model} & \textbf{pass@1} & \textbf{pass@2} & \textbf{pass@4} & \textbf{pass@8} & \textbf{pass@16} & \textbf{pass@32} & \textbf{pass@64} & \textbf{pass@128} \\
\midrule
\multicolumn{9}{c}{\textbf{IMO-AnswerBench}} \\
\midrule
Qwen2.5-7B-Instruct & 6.8 & 10.3 & 14.3 & 18.3 & 22.2 & 26.0 & 29.7 & 33.5 \\
GRPO & 7.7 & 11.2 & 15.0 & 18.6 & 22.3 & 25.9 & 29.6 & 32.8 \\
NuRL + GRPO & \textbf{8.0} & \textbf{11.4} & 14.8 & 18.2 & 21.3 & 24.4 & 27.6 & 30.8 \\
GRPO (Deepscaler) & 7.7 & 10.3 & 13.2 & 16.3 & 19.3 & 21.9 & 24.6 & 27.8 \\
\rowcolor[HTML]{DBEAF6} DAIL (Ours) & 6.4 & 10.3 & \textbf{15.1} & \textbf{20.3} & \textbf{25.3} & \textbf{29.9} & \textbf{34.1} & \textbf{37.8} \\
\midrule
\multicolumn{9}{c}{\textbf{BeyondAIME}} \\
\midrule
Qwen2.5-7B-Instruct & 4.1 & 6.0 & 8.7 & 12.1 & 16.6 & 22.4 & 30.1 & 41.0 \\
GRPO & 4.2 & 6.4 & 9.4 & 13.2 & 18.0 & 23.9 & 31.0 & 40.0 \\
NuRL + GRPO & 3.3 & 5.4 & 8.0 & 11.1 & 15.1 & 20.2 & 27.0 & 36.0 \\
GRPO (Deepscaler) & \textbf{4.9} & \textbf{7.1} & 9.6 & 12.4 & 15.5 & 19.0 & 22.9 & 27.0 \\
\rowcolor[HTML]{DBEAF6} DAIL (Ours) & 4.3 & 6.8 & \textbf{10.4} & \textbf{15.5} & \textbf{21.9} & \textbf{29.6} & \textbf{38.0} & \textbf{46.0} \\
\midrule
\multicolumn{9}{c}{\textbf{AIME 2024 / 2025}} \\
\midrule
Qwen2.5-7B-Instruct & 9.8 & 14.3 & 19.2 & 24.4 & 29.8 & 36.1 & 43.7 & 51.7 \\
GRPO & 9.0 & 13.2 & 17.6 & 22.6 & 29.1 & 37.1 & 45.4 & 53.3 \\
NuRL + GRPO & 11.0 & \textbf{16.3} & \textbf{21.5} & \textbf{26.4} & 31.5 & 36.8 & 42.0 & 46.7 \\
GRPO (Deepscaler) & \textbf{12.0} & 15.8 & 19.3 & 23.4 & 28.7 & 34.5 & 39.0 & 41.7 \\
\rowcolor[HTML]{DBEAF6} DAIL (Ours) & 8.8 & 13.4 & 18.8 & 25.2 & \textbf{33.4} & \textbf{43.5} & \textbf{54.1} & \textbf{63.3} \\
\bottomrule
\end{tabular}
\label{tab:pass_at_k_exact_numerical}
\end{table*}

\subsection{Exploratory Experiments and Connections to On-Policy Methods.}
During exploratory experiments, we experimented with an on-policy variant of DAIL, similar in mechanism to concurrent self-distillation methods~\cite{zhao2026self, hubotter2026reinforcement, shenfeld2026self}. However, we found that in this setting, loss converged extremely slowly, requiring many epochs on our intentionally small dataset. We speculate that this discrepancy is due to the difficulty of the questions used in our setting. As we note in \S\ref{subsec:learning}, training off-policy enables more efficient updates and parallel dataset generation, but likely sacrifices some performance gains~\cite{shao2024deepseekmath}. We leave such an investigation for future work.

\subsection{Other Evaluation Details and Ablations}
\label{appendix:eval_details_ablations}

\begin{figure}[htbp]
  \centering
  \begin{minipage}[b]{0.4\linewidth}
    \centering
    \includegraphics[width=\linewidth]{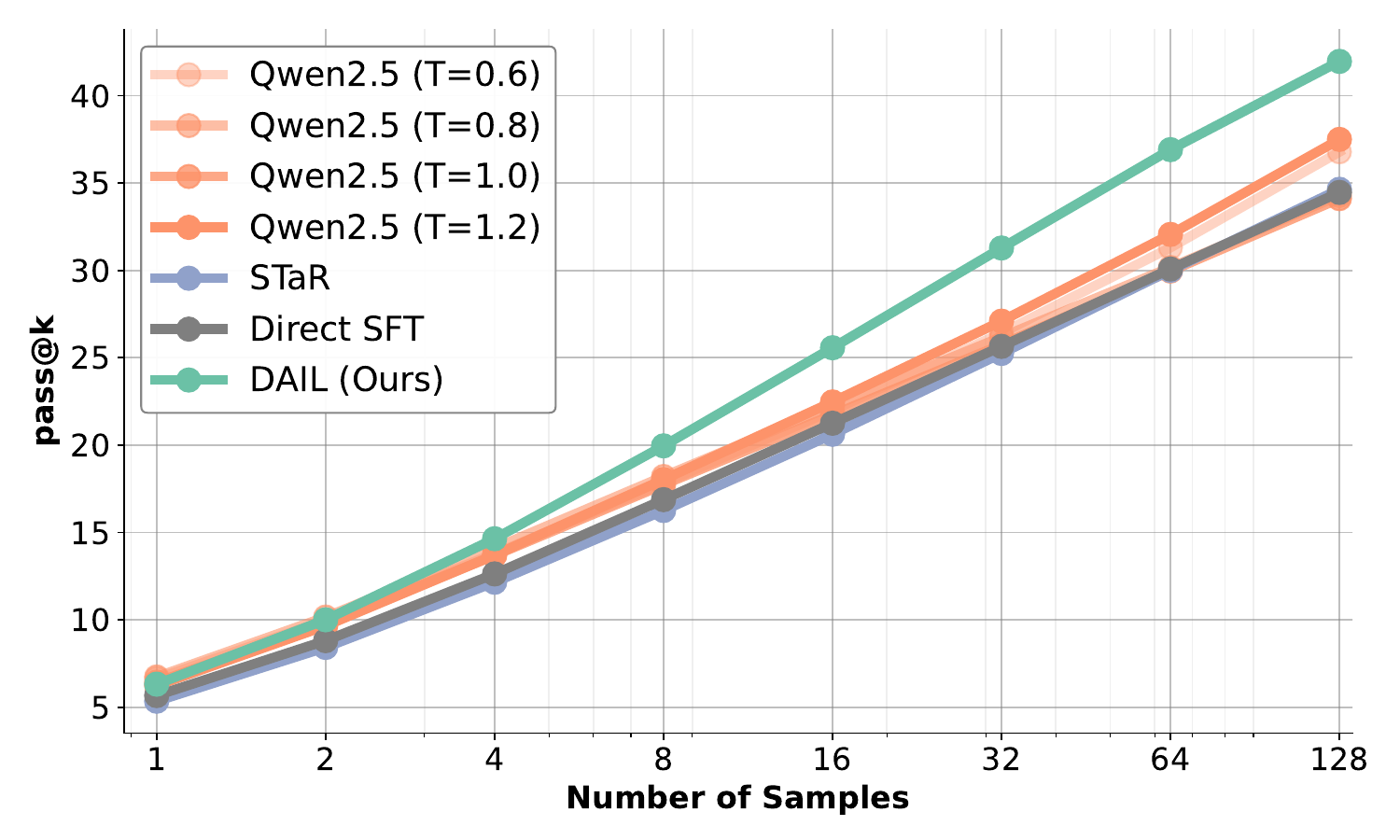}
    \caption{A version of Figure~\ref{fig:naive_baselines} for all $k$ values. These results include the aggregated performance of Qwen2.5-7B-Instruct when training on \texttt{e1-verifiable}.}
    \label{fig:all_naive_baselines}
  \end{minipage}
  \hfill
  \begin{minipage}[b]{0.5\linewidth}
    \centering
    \captionof{table}{Ablation of contrastive objective on BeyondAIME. Results are in the mixed policy rollouts setting. Contrastive loss outperforms only minimizing KL between the student and teacher.}
    \label{tab:kl_ablation}
    \begin{tabular}{@{}lc@{}}
    \toprule
    \textbf{Loss} & \textbf{pass@128} \\ \midrule
    Contrastive (Ours) & \textbf{46.0} \\
    $D_{\text{KL}}\big(M_\theta(\cdot|x, r_{<t}) \parallel M_\text{T}(\cdot|r_{<t})\big)$               & 44.0 \\
    Qwen2.5-7B-Instruct                 & 41.0 \\
    $D_{\text{KL}}\big(M_\theta(\cdot|x, r_{<t}) \parallel M_\text{NR}(\cdot|r_{<t})\big)$                & 38.1 \\ \bottomrule
    \end{tabular}
    \vspace{30pt}
  \end{minipage}
\end{figure}

\paragraph{Exact pass@$k$ results.} Exact numerical pass@$k$ results for Qwen2.5-7B-Instruct experiments can be found in Table~\ref{tab:pass_at_k_exact_numerical}.
\paragraph{Naive baselines.}
Figure~\ref{fig:all_naive_baselines} presents the version of Figure~\ref{fig:naive_baselines}, i.e., the aggregated performance of training Qwen2.5-7B-Instruct on \texttt{e1-verifiable} across benchmarks, for all values of $k$. 

\paragraph{Effect of contrastive objective.} Table~\ref{tab:kl_ablation} presents the performance of various ablations of DAIL's contrastive objective on BeyondAIME when using mixed policy rollouts. We find that the contrastive objective outperforms optimizing only $D_{\text{KL}}\big(M_\theta(\cdot|x, r_{<t}) \parallel M_\text{T}(\cdot|r_{<t})\big)$. Additionally, minimizing the divergence between the student and the negative reference, i.e., optimizing $D_{\text{KL}}\big(M_\theta(\cdot|x, r_{<t}) \parallel M_\text{NR}(\cdot|r_{<t})\big)$ degrades performance compared to the untrained model at $k = 128$. These results match the motivation for this objective and negative reference outlined in \S\ref{subsec:learning}.

\begin{figure}[h!]
    \centering
    \includegraphics[width=\linewidth]{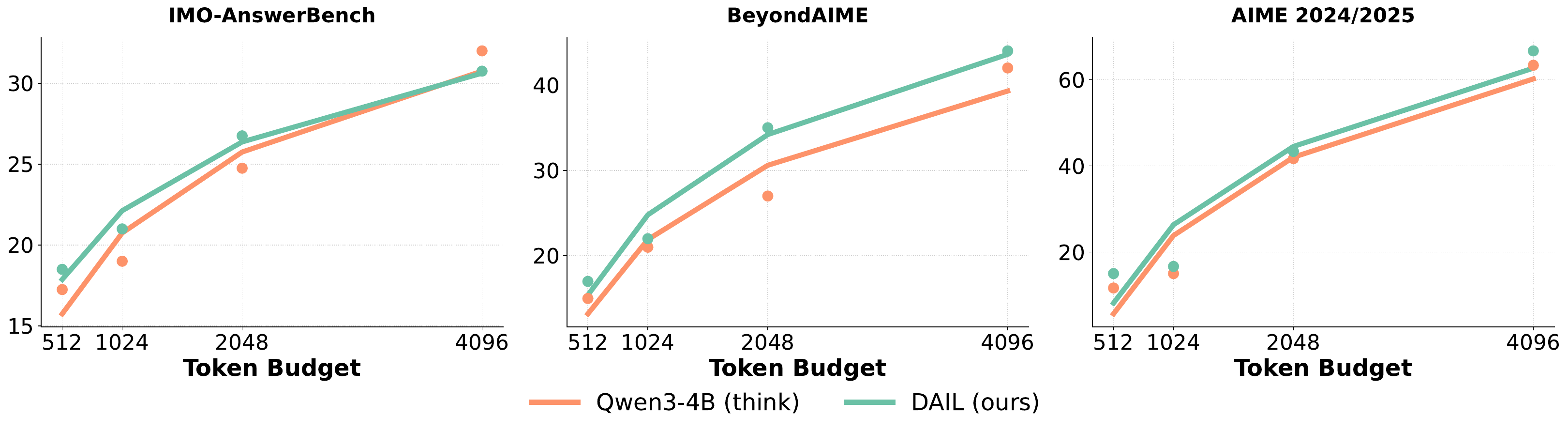}
    \includegraphics[width=\linewidth]{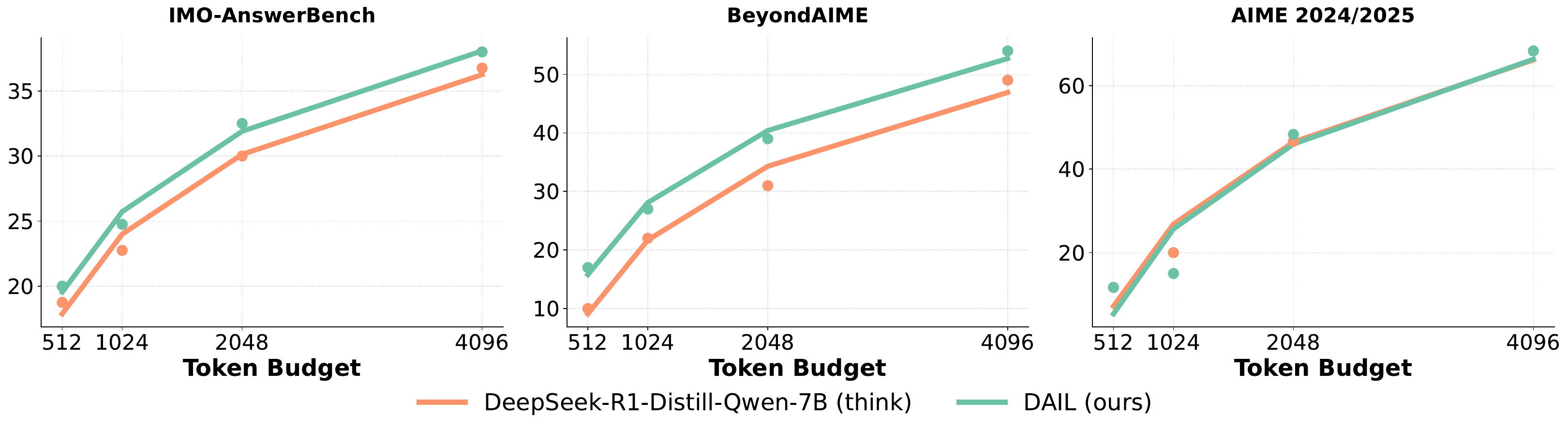}
    \caption{Test-time efficiency via coverage (pass@128) on smaller, less capable models. We find that for these less capable models, improvements are more modest, indicating that we need a baseline level of reasoning performance to apply DAIL for maximum improvement.}
    \label{fig:model_scale_tts_small}
\end{figure}

\paragraph{Comparing to standard distillation.} We also compare to standard supervised-distillation from Qwen3-235-A22B~\cite{yang2025qwen3} using the same number of tokens per example as DAIL. The results in Table~\ref{tab:supervised_distillation} demonstrate that both methods enable similar levels of coverage, making DAIL an attractive option as it does not require the use of a larger model for distillation.

\paragraph{Competent models are necessary for consistent improvement with DAIL.} Figure~\ref{fig:model_scale_tts_small} presents test-time scaling performance on smaller reasoning models, for which we find that the gains in performance are modest compared to the larger 8B and 14B models. Additionally, Table~\ref{tab:dail_limitations} presents BeyondAIME results on Llama-3.1-8B-Instruct, a much weaker model on mathematics tasks compared to the others used in our experiments. As shown, DAIL on \texttt{e1-verifiable} shows no improvement and in fact demonstrates a degradation in performance. This result demonstrates the limits of DAIL's ability to improve reasoning performance and the need for future work to understand the interaction between expert datasets and model strength.

\begin{table}[t]
\centering
\caption{Comparing DAIL to Distillation on Qwen (pass@128)}
\begin{tabular}{lcc}
\toprule
Token Limit & DAIL & Distillation from Qwen3-235B-A22B (think) \\
\midrule
512  & \textbf{18.0} & 15.0 \\
1024 & 23.0 & \textbf{24.0} \\
2048 & 34.0 & \textbf{37.0} \\
4096 & 49.0 & \textbf{55.0} \\
\bottomrule
\end{tabular}
\label{tab:supervised_distillation}
\end{table}

\begin{table}[t]
\centering
\caption{Llama-3.1-8B-Instruct DAIL performance on BeyondAIME. Results demonstrate that weaker models may struggle to learn with DAIL if the difficulty of training problems is far outside their capacity.}
\begin{tabular}{lcc}
\toprule
Model & pass@1 & pass@128 \\
\midrule
Llama-3.1-8B-Instruct & \textbf{0.64} & 24.0 \\
DAIL & 0.47 & 24.0 \\
\bottomrule
\end{tabular}
\label{tab:dail_limitations}
\end{table}

\begin{table}[h!]
\centering
\caption{Performance of DeepSeek-R1-Distill-Llama-8B on BeyondAIME.}
\begin{tabular}{lcc}
\toprule
Token Limit & DeepSeek-R1-Distill-Llama-8B & DAIL \\
\midrule
512  & 9.0  & \textbf{13.0} \\
1024 & 15.0 & \textbf{29.0} \\
2048 & 31.0 & \textbf{39.0} \\
4096 & 45.0 & \textbf{57.0} \\
\bottomrule
\end{tabular}
\label{tab:llama_backbone}
\end{table}

\paragraph{Performance on other backbones.} Following \citet{shao2025spurious}'s recommendation to perform experiments on different backbones besides Qwen to understand whether a method generalizes beyond a specific model family, we also apply DAIL to a distilled Llama DeepSeek-R1 variant. The results in Table~\ref{tab:llama_backbone} demonstrate that DAIL consistently improves performance on the Llama-based model.

\end{document}